\documentclass{article}
\usepackage{amssymb}
\usepackage{wrapfig}
\usepackage{graphicx}
\usepackage{amsmath}
\usepackage{booktabs}
\usepackage{makecell}
\usepackage{subcaption}
\usepackage[table]{xcolor}
\usepackage{float}

    \PassOptionsToPackage{numbers, compress}{natbib}
\usepackage[preprint]{neurips_2025}

\usepackage[utf8]{inputenc} % allow utf-8 input
\usepackage[T1]{fontenc}    % use 8-bit T1 fonts
\usepackage{hyperref}       % hyperlinks
\usepackage{url}            % simple URL typesetting
\usepackage{booktabs}       % professional-quality tables
\usepackage{amsfonts}       % blackboard math symbols
\usepackage{nicefrac}       % compact symbols for 1/2, etc.
\usepackage{microtype}      % microtypography
\usepackage{xcolor}         % colors
\usepackage{enumitem}

\title{OSVI-WM: One-Shot Visual Imitation for \\ Unseen Tasks using \\ World-Model-Guided  Trajectory Generation}

\author{%
  Raktim Gautam Goswami\textsuperscript{1}\thanks{This paper is supported in part by the Army Research Office under grant number W911NF-21-1-0155 and by the New York University Abu Dhabi (NYUAD) Center for Artificial Intelligence and Robotics, funded by Tamkeen under the NYUAD Research Institute Award CG010.}  
  \, Prashanth Krishnamurthy\textsuperscript{1} 
  \, Yann LeCun\textsuperscript{2,3} 
  \, Farshad Khorrami\textsuperscript{1} \\ [2mm]
  \textsuperscript{1}New York University Tandon School of Engineering \\
  \textsuperscript{2}New York University Courant Institute of Mathematical Sciences \,
  \textsuperscript{3}Meta-FAIR
}

\begin{document}

\maketitle

\begin{abstract}
  Visual imitation learning enables robotic agents to acquire skills by observing expert demonstration videos. In the one-shot setting, the agent generates a policy after observing a single expert demonstration without additional fine-tuning. 
  Existing approaches typically train and evaluate on the same set of tasks, varying only object configurations, and struggle to generalize to unseen tasks with different semantic or structural requirements. 
  While some recent methods attempt to address this, they exhibit low success rates on \textit{hard} test tasks that, despite being visually similar to some training tasks, differ in context and require distinct responses.
  Additionally, most existing methods lack an explicit model of environment dynamics, limiting their ability to reason about future states.
  To address these limitations, we propose a novel framework for one-shot visual imitation learning via world-model-guided trajectory generation. Given an expert demonstration video and the agent’s initial observation, our method leverages a learned world model to predict a sequence of latent states and actions. This latent trajectory is then decoded into physical waypoints that guide the agent’s execution. Our method is evaluated on two simulated benchmarks and three real-world robotic platforms, where it consistently outperforms prior approaches, with over 30\% improvement in some cases. The code is available at \href{https://github.com/raktimgg/osvi-wm}{https://github.com/raktimgg/osvi-wm}.
\end{abstract}

\section{Introduction}
\label{sec:introduction}
\begin{quote}
\vspace*{-0.4cm}
\centering
\textit{Imitation is the sincerest of flattery.} \\
\hfill --- Charles Caleb Colton\footnote{\textit{Lacon: Or Many Things in Few Words}, 1820.}
\vspace*{-0.3cm}
\end{quote}

Intelligent beings like humans are capable of learning a wide range of skills by observing and imitating others~\cite{vogt2007visuo}. Even toddlers, with a basic understanding of the world’s dynamics, solve complex tasks by watching \textit{expert demonstrations}. Inspired by this natural learning process, principles have been applied in robotics~\cite{vi_survey1, vi_survey2, vi_survey3, vi_survey4} where expert demonstrations, typically collected through human teleoperation, are used to teach agents to perform tasks through imitation. Imitation learning finds uses across diverse domains, including medical robotics, collaborative robotics, and industrial automation.

In a successful imitation, the agent discovers the skill demonstrated by the expert, adapts it to its own embodiment, and executes it within its environment~\cite{xskill}. In the one-shot visual imitation (OSVI) setting, the agent must derive a policy from a single expert demonstration video, without any additional training~\cite{osvi}. 
Notably, the expert and agent may differ in embodiment~\cite{tosvi}; in fact, both prior work~\cite{mimicplay, masked_visual_pre_training, structured_world_models} and our approach have utilized human demonstration videos as input.
Existing methods, however, often rely on the strong assumption that the training and testing tasks are nearly identical, typically differing only in object locations. As a result, generalization to semantically or structurally different tasks remains limited. MOSAIC~\cite{mosaic} and AWDA~\cite{awda} were among the first to investigate generalization to unseen tasks. In particular, AWDA~\cite{awda} demonstrated that the performance of standard methods like~\cite{tosvi, daml} degrades severely when evaluated on tasks that differ even marginally from those seen during training.
Although AWDA outperforms earlier methods~\cite{tosvi, daml}, it exhibits low success rates on test tasks that are visually similar but contextually different, such as opening vs. closing a sliding window.
Moreover, none of these approaches incorporates an explicit model of the environment's dynamics, limiting their ability to reason about long-term consequences or plan into the future.

\begin{figure}
    \centering    \includegraphics[width=.95\linewidth]{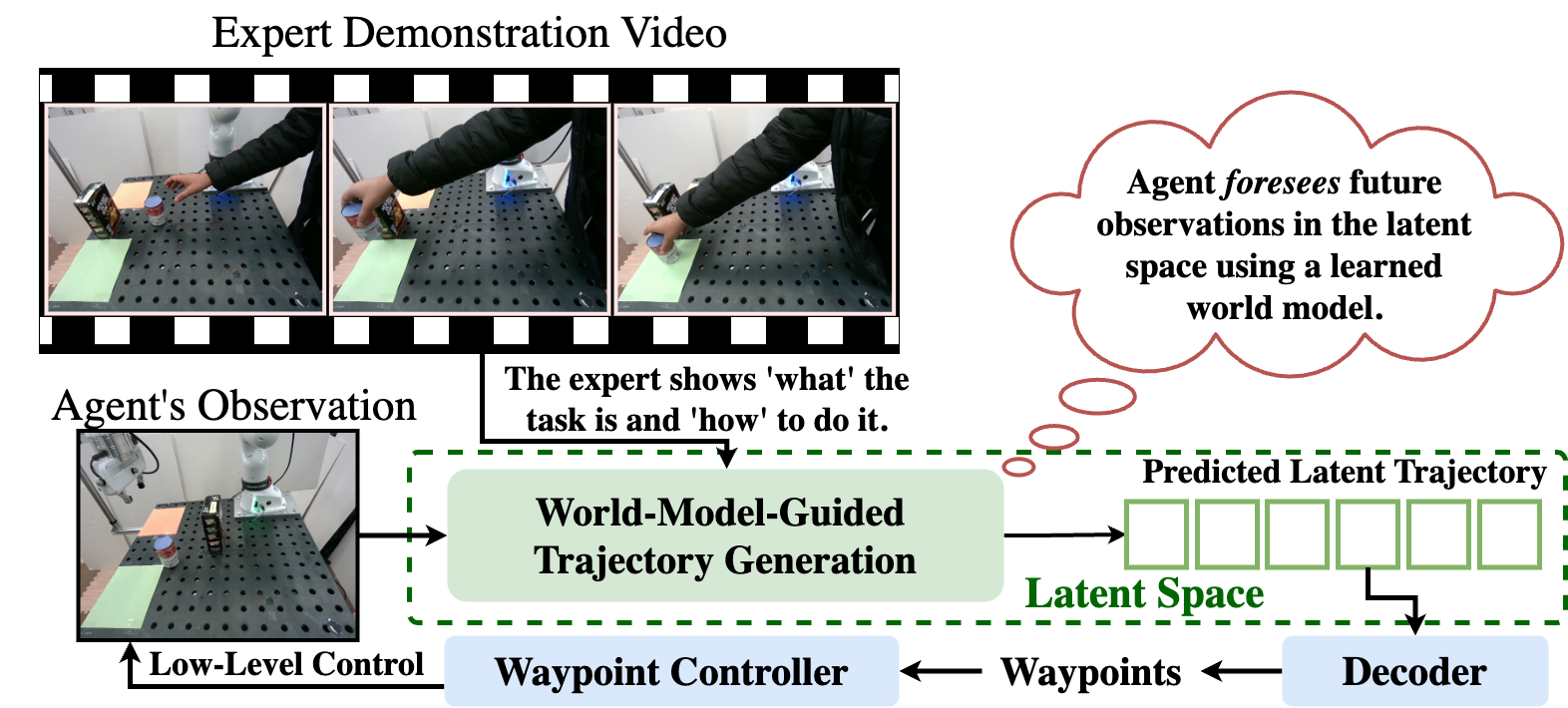}
    \caption{OSVI-WM infers the task from the expert demonstration and, along with the agent's observation ``foresees'' future latent states using a world-model-guided trajectory generation module. The predicted trajectory is decoded into physical waypoints for control.}
    \label{fig:teaser}
    \vspace*{-0.5cm}
\end{figure}

Mathematically modeling the environment using raw camera images is challenging due to the presence of diverse objects and their complex interactions. To address this, neural networks have been used to learn world models that predict how the environment will evolve in response to the agent's actions. These models support applications in reinforcement learning~\cite{rl_book, dreamer3, wmrl, iris, transdreamer, td_mpc2, daydreamer} and behavior cloning~\cite{dynamo, structured_world_models}.
Inspired by such advances, we propose \textit{OSVI-WM} (Fig.~\ref{fig:teaser}), a framework for one-shot visual imitation for unseen tasks using world-model-guided trajectory generation. Given an expert demonstration video and the agent’s initial observation, OSVI-WM encodes each image into a shared latent space representing the environment state for the respective image. These embeddings are recursively processed through learnable action and world models to predict a trajectory of future latent states, allowing the agent to plan into the future. Predicting actions in the latent space also helps mitigate the multi-modality of real-world action distributions.
The predicted latent trajectory is then decoded into physical waypoints~\cite{awda} using pooling and Multi-Layer Perceptron (MLP) layers. 

The predicted waypoints guide the agent’s execution and provide supervision for end-to-end training. To enhance the world model’s ability to capture environment dynamics, an auxiliary loss is applied on the predicted latent trajectory during training.
Further, OSVI-WM supports re-planning: if the agent doesn't accurately complete the task by following the predicted waypoints, it takes a new observation and plans a fresh set of waypoints based on its current state.
Notably, unlike methods that rely on large-scale pretraining or extensive demonstration data~\cite{structured_world_models, large_data1, large_data2}, OSVI-WM is trained directly within the task domain. 
In summary, our main contributions are:
\begin{itemize}[leftmargin=*, itemsep=0.1pt]
    \item An efficient end-to-end imitation learning architecture trained solely on in-domain data, without requiring large-scale pretraining.
    \item A novel world-model-guided trajectory generation module tailored for OSVI on unseen tasks.
    \item Robustness enhancement at test time by using a waypoint controller with re-planning.
    \item Extensive experiments in both simulated and real-world settings, demonstrating that OSVI-WM outperforms existing methods on unseen tasks.
\end{itemize}

% In summary, our main contributions are: (a) a novel world-model-based trajectory synthesis framework tailored for one-shot visual imitation for unseen tasks; (b) an efficient end-to-end imitation learning architecture trained solely on in-domain data, without requiring large-scale pretraining; (c) an attributed waypoint controller with a re-planning mechanism to enhance robustness at test time; and (d) extensive experiments in both simulated and real-world settings, demonstrating that OSVI-WM outperforms existing methods on unseen tasks.

% The rest of the paper is organized as follows: Sec. \ref{sec:related_works} covers related work, Sec. \ref{sec:methodology} presents the problem and our framework, Sec. \ref{sec:results} shows experimental results, and Sec. \ref{sec:conclusion} concludes the paper.
%===============================================================================

\section{Related Works}
\label{sec:related_works}
\noindent \textbf{One-Shot Imitation Learning}: 
Imitation learning, or learning from demonstrations, is broadly categorized into inverse reinforcement learning~\cite{irl1, irl2} and behavior cloning~\cite{bet, vqbet, cbet, baku, zsvi}. While the field has been extensively studied for years, a comprehensive review is beyond the scope of this paper; we, therefore, refer readers to existing surveys~\cite{vi_survey1, vi_survey3, vi_survey4}. Our method falls under behavior cloning, specifically in the one-shot visual imitation (OSVI) setting.
In standard behavior cloning, expert demonstrations and corresponding expert actions are available to train the agent. In contrast, OSVI~\cite{osvi} provides only a video of the expert performing the task, without access to the underlying actions. One of the earliest works on OSVI~\cite{osvi} introduced the problem setting and proposed a soft-attention-based learning framework for block stacking. Since then, a number of methods~\cite{osvi_meta, tosvi, tosvi2, imop, bcz} have improved performance in this setting. While ego-centric setups (e.g., DINOBot~\cite{dinobot}) have been explored, we focus on methods using external camera views. One such work, T-OSVI~\cite{tosvi}, uses transformers~\cite{transformers} to model long-range dependencies in demonstrations, and like many other works~\cite{rhyme, mimicplay, daml}, it supports expert-agent embodiment mismatch.

While these approaches have shown promising results, they are typically evaluated on the same tasks that were used for training with minor variations, such as changes in object positions or quantities. MOSAIC~\cite{mosaic} proposed a more challenging setup with completely unseen test tasks, building on which, AWDA~\cite{awda} demonstrated that existing methods like T-OSVI~\cite{tosvi} and DAML~\cite{daml} perform poorly in such settings. AWDA improved generalization through attributed waypoints, demonstration augmentation, and image mixup, but achieved low success rates on \textit{hard} tasks requiring different responses despite visual similarity. While IMOP~\cite{imop} used point clouds to aid generalization, we focus solely on monocular RGB images, which are more accessible and require no specialized hardware. Notably, none of these approaches models environment dynamics, limiting their ability to reason about future states. OSVI-WM addresses this gap through a world-model-guided trajectory generation module that enables future-aware planning and improved generalization to unseen tasks.

\noindent \textbf{World Model in Robotics}: 
Intelligent beings like animals are believed to possess internal models of the world that support control and planning for task execution~\cite{human_wm1, human_wm2, human_wm3, human_wm4}. Inspired by this, researchers have increasingly explored the use of world models in robotics~\cite{robotic_wm_survey}. In reinforcement learning (RL), several works~\cite{rwm, wmrl, dreamer, dreamer2, dreamer3, planet, iris, daydreamer, transdreamer, transformer_wm, deep_visual_foresight} have focused on learning world models from agent rollouts. These models are then used as environments for RL policy training, often also predicting rewards based on the agent’s actions.
Most prior works train world models directly in the pixel space, i.e., conditioned on the current image and action, the model predicts the next-frame image, often also integrated with diffusion-based frameworks.
These world models, however, requires modeling low-level pixel details that are often irrelevant for downstream tasks~\cite{jepa}. To address this, recent methods~\cite{td_mpc, td_mpc2, dawm, stochastic_latent_ac} learn world models in a latent embedding space, allowing for more efficient and abstract representations of the environment.

Beyond RL, world models have been explored for broader robotic learning tasks. For instance, SWIM~\cite{structured_world_models} pre-trains a world model from human videos and fine-tunes it on robot data in an unsupervised manner. GR1~\cite{gr1} and~\cite{spatiotemporal_pre_training} propose pre-training strategies to improve downstream robotic motor control. Most of these approaches, however, rely on large-scale internet data or massive pretraining datasets, leading to high computational demands.
DynaMo~\cite{dynamo}, in contrast, proposed in-domain latent dynamics pretraining for the encoder, which was used to train downstream policies. Similarly, DINO-WM~\cite{dino_wm} used the pre-trained DINOv2~\cite{dinov2} model (with frozen weights) to build in-context world models and applied model predictive control based optimization for task execution.
Inspired by these advances, our proposed framework, OSVI-WM, learns a world model entirely from in-domain data. It predicts latent trajectories based on an expert demonstration and the agent’s current observation. Unlike prior work, OSVI-WM, including its world model, is trained end-to-end, without any pretraining, simplifying the pipeline and reducing computational complexity.

\section{Our Method}
\label{sec:methodology}
\begin{figure}
    \centering
    \includegraphics[width=.95\linewidth]{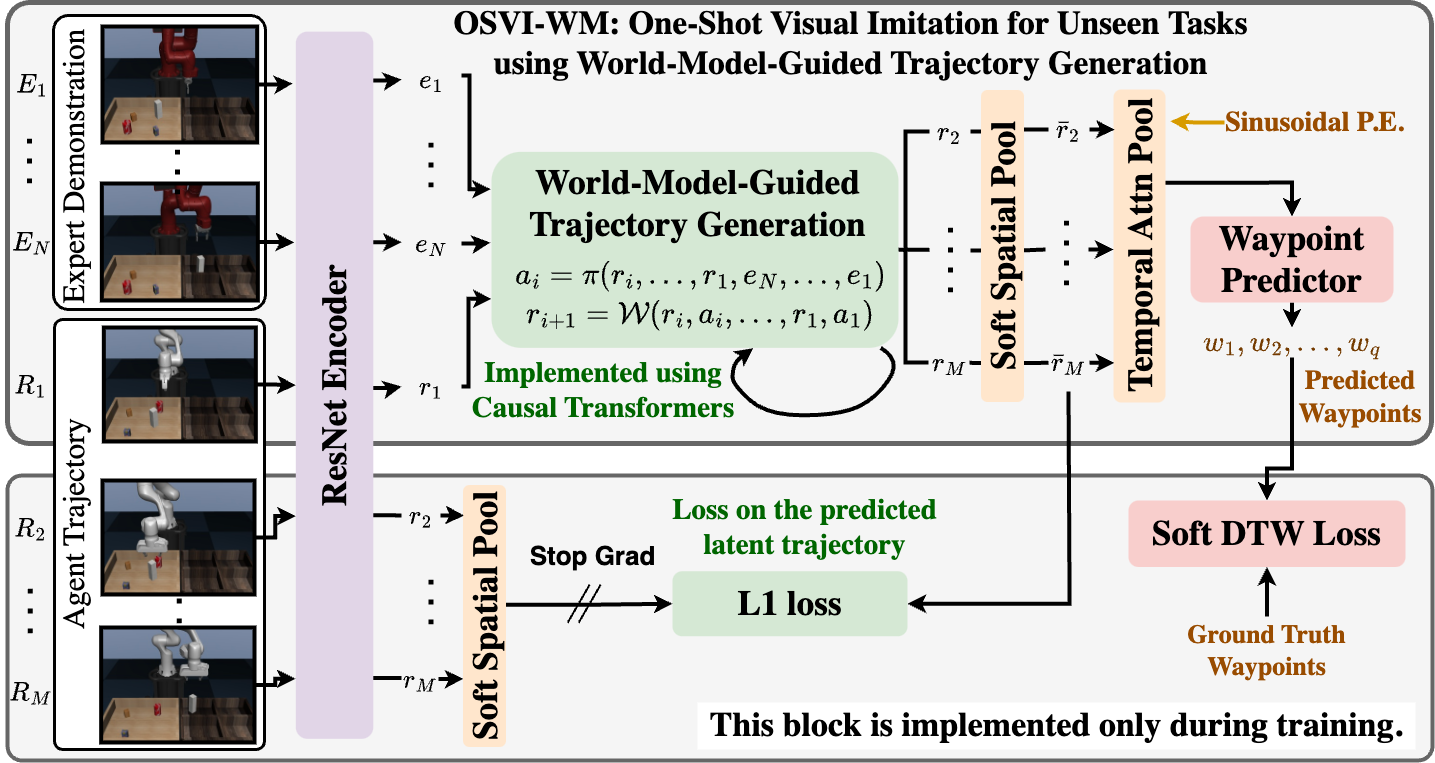}
    \caption{\textbf{OSVI-WM}: The expert demonstration frames $E_1,\dots,E_N$ and the agent’s current observation $R_1$ are encoded into a latent space using a ResNet encoder. A world-model-guided trajectory generation module predicts future latent states $r_2,\dots,r_M$, which are decoded into physical waypoints for control. During training, supervision is applied on the predicted waypoints and latent states.}
    \label{fig:flow}
    \vspace*{-0.6cm}
\end{figure}

\noindent \textbf{Problem Formulation}: 
We formulate the problem following prior work in OSVI~\cite{tosvi, mosaic, awda}. 
Let $\mathcal{T}=\{\mathcal{T}_1, \dots, \mathcal{T}_K\}$ denote a set of tasks, partitioned into disjoint training ($\mathcal{T}_{train}$) and test ($\mathcal{T}_{test}$) sets. In contrast to~\cite{tosvi} and consistent with~\cite{awda}, our setup uses different training and test tasks, rather than minor variations of the same task.
Each training task consists of multiple sequences of expert and agent trajectories. The expert trajectory comprises demonstration video frames $\{E_1, \dots, E_N\}$, while the agent trajectories are represented by observation-state pairs $(\mathcal{R}, \mathcal{S})$ where $\mathcal{R} = \{R_1, \dots, R_M\}$ denotes the observation frames and $\mathcal{S}$ denotes the corresponding agent states.
Within each task, all expert and agent trajectories correspond to variations (e.g., different object configurations) of the same high-level task.
The model, trained on $\mathcal{T}_{train}$, is evaluated on $\mathcal{T}_{test}$, which contains only expert demonstration videos (no agent rollouts). The performance is measured by recording the success rate of the agent in completing these test tasks. The key challenges in this setup are: (a) test tasks $\mathcal{T}_{test}$ differ from the training tasks $\mathcal{T}_{train}$; (b) the expert and the agent may have different embodiments; (c) there is no direct alignment between expert and agent states or actions; and (d) to avoid reliance on costly large-scale data, the method should be trained in-domain.

\noindent \textbf{Method Overview}: Given expert demonstration video frames $E_1, \dots, E_N$ and the agent’s initial observation $R_1$, OSVI-WM (Fig.~\ref{fig:flow}) encodes them using a shared ResNet-18~\cite{resnet} encoder to obtain latent states for the expert and agent. These are passed to the world-model-guided trajectory generation module (Sec.~\ref{sec:wm}) to predict a trajectory of future latent states, which is pooled, first spatially, then temporally, into a compact representation. Using this as input, an MLP-based waypoint predictor generates physical waypoints (Sec.~\ref{sec:waypoint_prediction}). Loss functions are applied to the predicted waypoints and the latent states predicted by the world model (Sec.~\ref{sec:training_settings}).

\subsection{World-Model-Guided Trajectory Generation}
\label{sec:wm}
\begin{wrapfigure}{r}{0.4\columnwidth}
    \vspace*{-0.8cm}
    \centering
    \includegraphics[trim={0 20 0 20},clip,width=0.36\columnwidth]{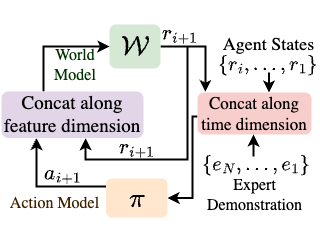}
    \caption{\textbf{WM-Guided Trajectory Generation}: Starting with the agent’s initial observation and the expert demonstration, future states are recursively predicted using action and world models.}
    \label{fig:wm}
    \vspace*{-0.3cm}
\end{wrapfigure}
The expert demonstration and the agent’s initial observation frames are each processed through a ResNet-18~\cite{resnet} encoder, producing feature maps of shape $(F, H, W)$ before the encoder's global pooling layer, where $F = 512$ and $(H, W)$ depend on the input resolution. These features are flattened to produce latent states for the expert frames ($e_1, \dots, e_N$) and the agent's observation ($r_1$), each frame with shape $(F \times H \times W)$. A latent state is a compact, abstract representation of the environment that encodes key features such as the configuration of objects and dynamic elements. A latent action similarly represents an abstract transformation that drives state transitions. Operating in this latent space further helps mitigate the multi-modality of real-world action distributions.
The world-model-guided trajectory generation module consists of an action model $\pi$ and world model $\mathcal{W}$. At timestep $i$, $\pi$ predicts latent action $a_i$ using the agent's current and past latent states along with the expert demonstration's latent states. This action is used by $\mathcal{W}$ to predict the agent's next latent state:
\begin{align}
    a_{i} = \pi(r_{i},\dots, r_1, e_N, \dots, e_1), &\quad r_{i+1} = \mathcal{W}(r_{i}, a_{i},\dots, r_1, a_1); \hspace{0.5cm}i=\{1,\dots,M-1\}.
\end{align}
% \begin{equation}
% \begin{aligned}
% a_1 = \pi(r_1, e_N, \dots, e_1), &\quad r_2 = \mathcal{F}(r_1, a_1) \\
% a_2 = \pi(r_2, r_1, e_N, \dots, e_1), &\quad r_3 = \mathcal{F}(r_2, a_2, r_1, a_1) \\
% &\vdots \\
% a_{M-1} = \pi(r_{M-1},\dots, r_1, e_N, \dots, e_1), &\quad r_M = \mathcal{F}(r_{M-1}, a_{M-1},\dots, r_1, a_1) \\
% \end{aligned}
% \label{eq:trajectory_synthesis}
% \end{equation}
This multi-step recursive process (Fig.~\ref{fig:wm}) generates all future latent states of the agent $(r_2, \dots, r_M)$, forming the generated trajectory. The world model (WM) essentially learns the environment’s dynamics to predict the next latent state based on the current latent state and latent action.

Within the action model $\pi$, at each step $i$, the expert states $\{e_N, \dots, e_1\}$ and agent states $\{r_i, \dots, r_1\}$ are concatenated along the temporal dimension, resulting in shape $(N+i, F \times H \times W)$. This is processed through causal transformer blocks, and the output corresponding to the final temporal index is used as the latent action $a_i$.
For the world model $\mathcal{W}$, the latent actions $\{a_i, \dots, a_1\}$ are concatenated with the corresponding agent states $\{r_i, \dots, r_1\}$ along the feature dimension to obtain shape $(i, 2F \times H \times W)$. This sequence is processed through a causal transformer, and the output at the last time step is projected to shape $(F \times H \times W)$ to produce the next state $r_{i+1}$.

\subsection{Predicting Waypoints}
\label{sec:waypoint_prediction}
\noindent \textbf{Spatial and Temporal Pooling}:
The predicted trajectory of future states (${r_2, \dots, r_M}$) is reshaped back to shape $(F, H, W)$ at each timestep to retain spatial structure. A spatial pooling module~\cite{tosvi} then applies a 2D softmax across each feature map and computes an expected 2D coordinate over a $[-1, 1]$ grid for each feature dimension. These coordinates are concatenated, producing a spatially pooled output of shape $2F$ for each of the $M-1$ timesteps ($\bar{r}_2,\dots,\bar{r}_{M}$). 
Sinusoidal positional embeddings are then added along the temporal dimension, and an attention pooling module with a learned query summarizes the sequence into a single $2F$-dimensional vector.

\noindent \textbf{Attributed Waypoint Prediction}: 
The pooled vector is passed through a two-layer MLP to predict 5 attributed waypoints, each of dimension 4. Following~\cite{awda}, each waypoint encodes the end-effector’s 3D position in the camera's coordinate frame and a binary gripper state (open/closed). These positions are transformed into the robot’s coordinate frame using a known camera-to-robot transformation. A waypoint controller, using inverse kinematics of the robot, computes low-level control commands for execution. To ensure accurate grasping, especially in the final centimeters before contact, we follow~\cite{awda} and use an end-effector-mounted depth camera to correct residual pose errors.

\noindent \textbf{Waypoint Re-Planning}:
To improve robustness during execution, OSVI-WM supports waypoint re-planning, which activates if the agent fails to complete the task using the predicted waypoints. In simulation, success or failure is automatically determined from environment configurations. 
Although we do not use re-planning on real-world benchmarks in this work, success (or completion) signals could be obtained via learned reward/value models~\cite{td_mpc} or vision–language-based completion detectors~\cite{vlm_for_completion1, vlm_for_completion2}. In OSVI-WM, once all waypoints are executed, the agent captures a new observation, which, combined with the original expert demonstration, generates updated waypoints. Notably OSVI-WM outperforms existing baselines even without re-planning (Fig.~\ref{fig:ablation_replanning}).

% \vspace*{-0.2cm}
\subsection{OSVI-WM Training}
\label{sec:training_settings}
\noindent \textbf{Loss Functions}:
We jointly optimize the world model and the waypoint predictions. To supervise the world model, ground-truth agent observations $R_2, \dots, R_M$ from the training set are encoded into latent states using our encoder, followed by the spatial pooling layer to produce ${\hat{r}_2, \dots, \hat{r}_M}$. An $L_1$ loss is applied between these and the predicted latent states after spatial pooling (${\bar{r}_2, \dots, \bar{r}_M}$): $\mathcal{L}_{wm} = \frac{1}{M-1}\sum_{i=2}^M\|\bar{r}_i-\hat{r}_i\|_1.$
To prevent model collapse and ensure stable training, we apply a stop-gradient to ${\hat{r}_2, \dots, \hat{r}_M}$, preventing backpropagation through the ground-truth latent states. For waypoint supervision, we interpolate the ground-truth agent trajectory into a denser sequence and use the differentiable soft dynamic time-warping loss~\cite{soft_dtw} ($\mathcal{L}_{sdtw}$). Although only 5 waypoints are used at inference, we follow~\cite{awda} and predict multiple sets (1 to 5 waypoints) during training, applying separate losses for each set to improve supervision across different trajectory granularities.
The overall training objective is $\mathcal{L} = \mathcal{L}_{sdtw} + \alpha(\tau)\mathcal{L}_{wm}$
where $\alpha(\tau)$ is a scheduling factor that balances the two loss terms based on the training iteration $\tau$.
For the Meta-World~\cite{metaworld} benchmark experiments in Sec.~\ref{sec:results}, $\alpha(\tau)$ is initialized to 1 and decays exponentially to 0.05 by the end of training. For Human-Franka-PP, $\alpha(\tau)=10$; for all other datasets, we keep $\alpha(\tau) = 1$.

\noindent \textbf{Training Settings}: 
Each expert demonstration is downsampled to $N = 10$ frames, and each agent trajectory to $M = 6$ frames. Following prior work~\cite{tosvi, awda}, we pair every expert demonstration with every agent trajectory within the same task, resulting in a combinatorial dataset expansion. We train for 10 epochs in simulation and 100 in real-world settings using AdamW~\cite{adamw} optimizer with a one-cycle learning rate scheduler~\cite{one_cycle_lr} (start: 0.0002, end: $2 \times 10^{-7}$), and a batch size of 128. We apply asymmetric demonstration image mixup~\cite{awda} regularization to help the model focus on task-relevant cues rather than video semantics. Although training runs for 10 epochs, we use early stopping for Meta-World to avoid overfitting and report results using the overall best-performing checkpoint. For simulation runs, we used an RTX A4000 GPU with 128 GB RAM and Intel i9 CPU, and for real-world, we used an RTX 2080 Ti, 64 GB RAM, and Intel i7 CPU.

\section{Experiments}
\begin{figure}
    \centering
    \includegraphics[trim={0 5 0 0},clip, width=\linewidth]{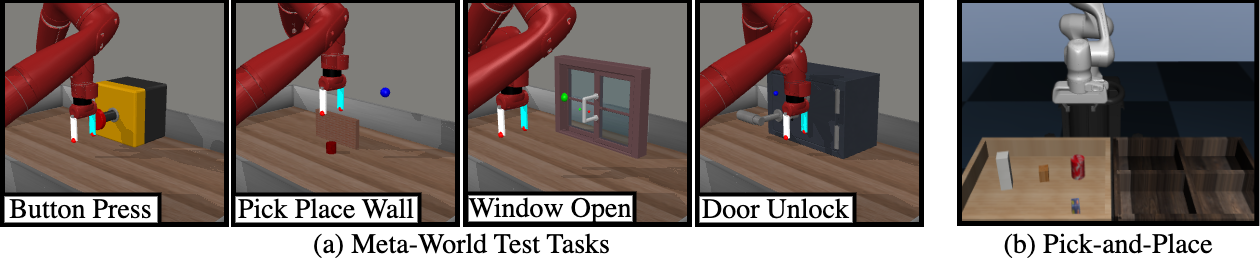}
    \caption{\textbf{Simulation Environments}: Test tasks are different from the ones used for training. Additionally,  Pick-and-Place uses different embodiments for expert and agent.}
    \label{fig:simulation_tasks}
    \vspace*{-0.2cm}
\end{figure}
\begin{figure}[t]
    \centering
    \includegraphics[width=\linewidth]{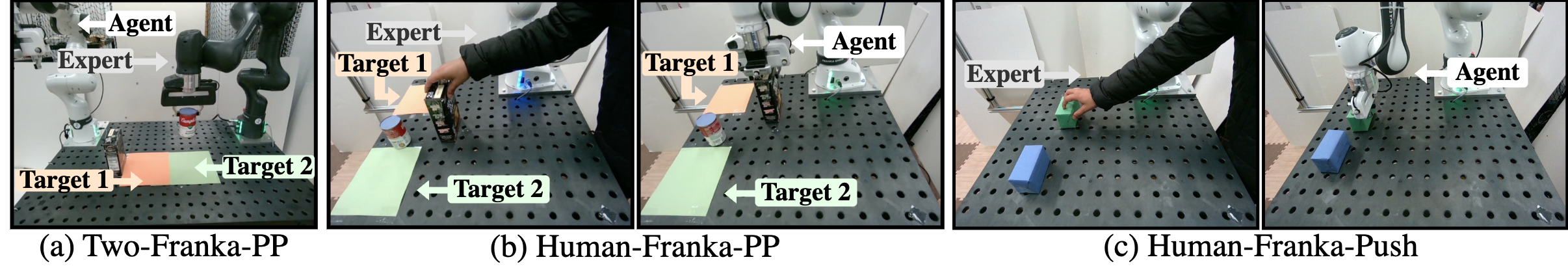}
    \caption{\textbf{Real-World Environments}: (a) Pick-and-place setup with expert (gray) and agent (white) Franka arms mounted at different locations. (b) Similar setup with a human expert and Franka agent. (c) Box push setup with human expert and Franka agent.}
    \label{fig:real_tasks}
    \vspace*{-0.5cm}
\end{figure}
\label{sec:results}
Our experiments are designed to address the following key questions:
(a) How well does OSVI-WM generalize to novel, unseen tasks?
(b) Can OSVI-WM be used for real-world robotic tasks?
(c) What is the impact of the world-model-guided trajectory generation module on overall performance?
(d) Which components of OSVI-WM are most critical to its success?

\subsection{Evaluation Settings}

% \begin{wrapfigure}{r}{0.4\columnwidth}
%     \vspace*{-0.5cm}
%     \centering
%     \includegraphics[trim={0 0 0 0},clip,width=0.38\columnwidth]{Figs/franka_crrl.png}
%     \caption{Real Robot Platform.}
%     \label{fig:franka_crrl}
%     % \vspace*{-0.5cm}
% \end{wrapfigure}

\textbf{Environments}: 
We evaluate OSVI-WM on simulation (Fig.~\ref{fig:simulation_tasks}) and real-world benchmarks (Fig.~\ref{fig:real_tasks}):
\begin{itemize}[itemsep=0.05pt]
    \item[(a)] \textbf{Meta-World}~\cite{metaworld}: A simulation benchmark of 50 manipulation tasks, split into 46 for training and 4 for testing following~\cite{awda}. Test tasks are different from train tasks and are divided into `easy' tasks (Button-Press-V2, Pick-Place-Wall-V2), which differ from training tasks due to the presence/absence of distractor, and `hard' tasks Window-Open-V2, Door-Unlock-V2, which, despite being visually similar to Window-Close-V2 and Door-Lock-V2 from the training set, respectively, differ in context and require distinct responses.
    \item[(b)] \textbf{Pick-and-Place}~\cite{tosvi}: A simulated benchmark involving 4 objects and 4 targets, yielding 16 tasks. Following~\cite{awda}, we use 14 tasks for training and 2 for testing. This benchmark features different embodiments: the expert uses a Sawyer arm, and the agent uses a Franka arm.
    \item[(c)] \textbf{Two-Franka-PP} (Fig.~\ref{fig:real_tasks}a): A real-world setup with two Franka arms mounted at different tabletop locations, with the gray arm as the expert and the white arm as the agent. The tasks are similar to Pick-and-Place above, with two objects and two targets (4 tasks total). We use 3 tasks for training and 1 for testing. The difference in mounting location between the expert and agent introduces additional imitation complexity.
    \item[(d)] \textbf{Human-Franka-PP} (Fig.~\ref{fig:real_tasks}b): Similar to Two-Franka-Env, but the expert is a human arm and the agent is a Franka robot. This greater embodiment mismatch increases task difficulty. As before, we use 3 tasks for training and 1 for testing.
    \item[(e)] \textbf{Human-Franka-Push} (Fig.~\ref{fig:real_tasks}c): This features two colored boxes on a tabletop. The task involves selecting a box and pushing it either forward or backward (four tasks total). We train on three and test on the held-out one. The expert is a human and agent is a Franka arm.
\end{itemize}

In the pick-and-place tasks above, sequences from test object-target pairs are excluded from training. While this may appear to be an extension of training tasks, most prior methods require training on all pairs to succeed. Generalizing to an unseen pair requires the model to truly follow the expert, rather than rely solely on scene semantics. Additional environmental details are in the appendix.

\noindent \textbf{Baselines}:
As noted in Sec.~\ref{sec:introduction} and Sec.~\ref{sec:related_works}, most one-shot visual imitation methods are evaluated on tasks that are minor variations of training tasks. AWDA~\cite{awda} is the only method that uses the same sensors and environment settings as ours and evaluates on unseen test tasks, making it our primary baseline. We also compare against two widely used methods: DAML~\cite{daml} and T-OSVI~\cite{tosvi}. In real-world experiments, T-OSVI fails to achieve successful grasps. To mitigate this, we augment it with the same depth-based gripper correction used in OSVI-WM and AWDA, where an end-effector-mounted depth camera corrects residual pose errors. We refer to this variant as T-OSVI* and report its results in Table~\ref{tab:real_robot_results}. For OSVI-WM, re-planning is used only on the MetaWorld `hard' tasks, as we observe strong performance even without it on all other benchmarks.

\begin{table}[t]
    \centering
    \setlength{\tabcolsep}{8pt}
    \caption{Success rates (in \%) comparison on the Meta-World~\cite{metaworld} and Pick-and-Place~\cite{tosvi} simulation benchmarks. Best results are highlighted. We also report if a method uses additional training data.}
    \vspace{0.1cm}
    \begin{tabular}{l c c  c c c}
    \toprule
    \textbf{Method} & \makecell[c]{\textbf{Add.} \textbf{Data}} & \makecell[c]{\textbf{Pick} \textbf{Place}} & \multicolumn{3}{c}{\textbf{Meta-World}} \\
    \cmidrule(lr){4-6}
    & & & \textbf{Easy} & \textbf{Hard} & \textbf{All} \\
    \hline
    DAML~\cite{daml}         & \cellcolor{green!20}No  & 1   & 4   & 8    & 6    \\
    T-OSVI~\cite{tosvi}      & \cellcolor{green!20}No  & 10  & 50  & 7    & 28.5 \\
    AWDA~\cite{awda}         & \cellcolor{green!20}No  & \cellcolor{gray!20}\textbf{100} & 74  & 11   & 42.5 \\
    AWDA~\cite{awda}         & \cellcolor{red!20}Yes   & \cellcolor{gray!20}\textbf{100} & 73  & 30   & 51.5 \\
    \textbf{OSVI-WM (Ours)}           & \cellcolor{green!20}No  & \cellcolor{gray!20}\textbf{100} & \cellcolor{gray!20}\textbf{96} & \cellcolor{gray!20}\textbf{71.5} & \cellcolor{gray!20}\textbf{83.8} \\
    \bottomrule
    \end{tabular}
    \label{tab:simulation_results}
    \vspace*{-0.5cm}
\end{table}

\vspace*{-0.2cm}
\subsection{Results}
\noindent \textbf{How well does OSVI-WM generalize to novel, unseen tasks?}
\begin{table}[t]
    \centering
    \caption{Real-World experiments: Success rates and execution breakdowns (in \%) are reported. T-OSVI* denotes T-OSVI~\cite{tosvi} aided with end-effector depth sensing for improved grasping.}
    \vspace{0.1cm}
    \setlength{\tabcolsep}{4pt}
    \renewcommand{\arraystretch}{0.8}
    \begin{tabular}{l c cc c cc c c}
    \toprule
    & \multicolumn{3}{c}{\textbf{Two-Franka-PP}} 
    & \multicolumn{3}{c}{\textbf{Human-Franka-PP}} & \multicolumn{2}{c}{\textbf{Human-Franka-Push}} \\
    \cmidrule(lr){2-4} \cmidrule(lr){5-7} \cmidrule(lr){8-9}
    \textbf{Method} 
    &  & \multicolumn{2}{c}{Breakdown} 
    & & \multicolumn{2}{c}{Breakdown}
    & & \multicolumn{1}{c}{Breakdown}\\
     \cmidrule(lr){3-4}  \cmidrule(lr){6-7} \cmidrule(lr){9-9}
    & \makecell[tc]{Overall\\Success} & \makecell[tc]{Reach\\ Obj.} & \makecell[tc]{Grasp\\ Obj.}
    & \makecell[tc]{Overall\\Success} & \makecell[tc]{Reach\\ Obj.} & \makecell[tc]{Grasp\\ Obj.}
    & \makecell[tc]{Overall\\Success} & \makecell[tc]{Reach\\ Obj.} \\
    \midrule
    T-OSVI~\cite{tosvi}     & 0& 52& 0& 0& 0& 0 & 16& 96\\
    T-OSVI*~\cite{tosvi}    & 4& 52& 36& 0& 0& 0 & -& -\\
    AWDA~\cite{awda}        & 88& 88& 88& \cellcolor{gray!20}\textbf{92}& 92& 92 & 76& 100\\
    \textbf{OSVI-WM (Ours)} & \cellcolor{gray!20}\textbf{96}& 96& 96& \cellcolor{gray!20}\textbf{92}& 92& 92 & \cellcolor{gray!20}\textbf{100}& 100\\
    \bottomrule
    \end{tabular}
    \label{tab:real_robot_results}
    \vspace*{-0.5cm}
\end{table}
We compare OSVI-WM against baseline methods in Table~\ref{tab:simulation_results} on the Pick-and-Place~\cite{tosvi} and Meta-World~\cite{metaworld} benchmarks. OSVI-WM is evaluated over 100 rollouts per task with varied object configurations, and success rates are reported. Baseline results are reported from~\cite{awda}. OSVI-WM outperforms prior methods, achieving success rates up to 30\% higher than AWDA even when AWDA uses additional training data. OSVI-WM's ability to predict future latent states enables effective planning, resulting in strong performance even on the Meta-World `hard' tasks. Failure cases in this setting often occurred in edge cases where objects, such as windows or doors, were positioned far from the agent, making manipulation challenging due to depth ambiguity in external RGB observations. Nonetheless, OSVI-WM achieves a 71.5\% success rate on these tasks, outperforming all baselines.

\noindent \textbf{Can OSVI-WM be used for real-world robotic tasks?}
We evaluate OSVI-WM and baselines on the real-world setups, each tested over 25 rollouts using the same 25 object configurations across methods for fair comparison (Table~\ref{tab:real_robot_results}). These setups introduce added complexity: in Two-Franka-PP, the expert and agent robots are mounted at different locations, while in Human-Franka-PP and Human-Franka-Push, the expert is a human. Despite these challenges, OSVI-WM achieves consistently high success rates, showing strong real-world applicability.
The table also breaks performance into sub-tasks like reaching and grasping. In Two-Franka-PP, T-OSVI~\cite{tosvi} reaches the object in 52\% of trials but fails to grasp it. T-OSVI* (i.e., T-OSVI with grasp correction) improves grasping to 36\% but completes the full task only once (4\%). AWDA succeeds in 88\% of the trials, failing due to incorrect object selection. OSVI-WM performs best, succeeding in 96\% of the trials.
Human-Franka-PP is more challenging due to the embodiment gap. In this setting, T-OSVI and T-OSVI* fail to reach the correct object, while AWDA and OSVI-WM both succeed in 92 \% of the trials. 
In Human-Franka-Push, all methods consistently reach the correct object. However, due to the low margin for error in this setting, AWDA and T-OSVI often collided with the object during fine manipulation, leading to reduced success rates of 76\% and 16\%, respectively. OSVI-WM executed these movements more accurately, achieving 100\% success. 
Since there is no grasping involved in this task, we do not run T-OSVI*.
% Overall, these results validate OSVI-WM’s robustness in real-world imitation tasks and its advantage over prior methods.

\begin{figure}
    \centering
    \includegraphics[trim={0 5 0 0},clip,width=.86\linewidth]{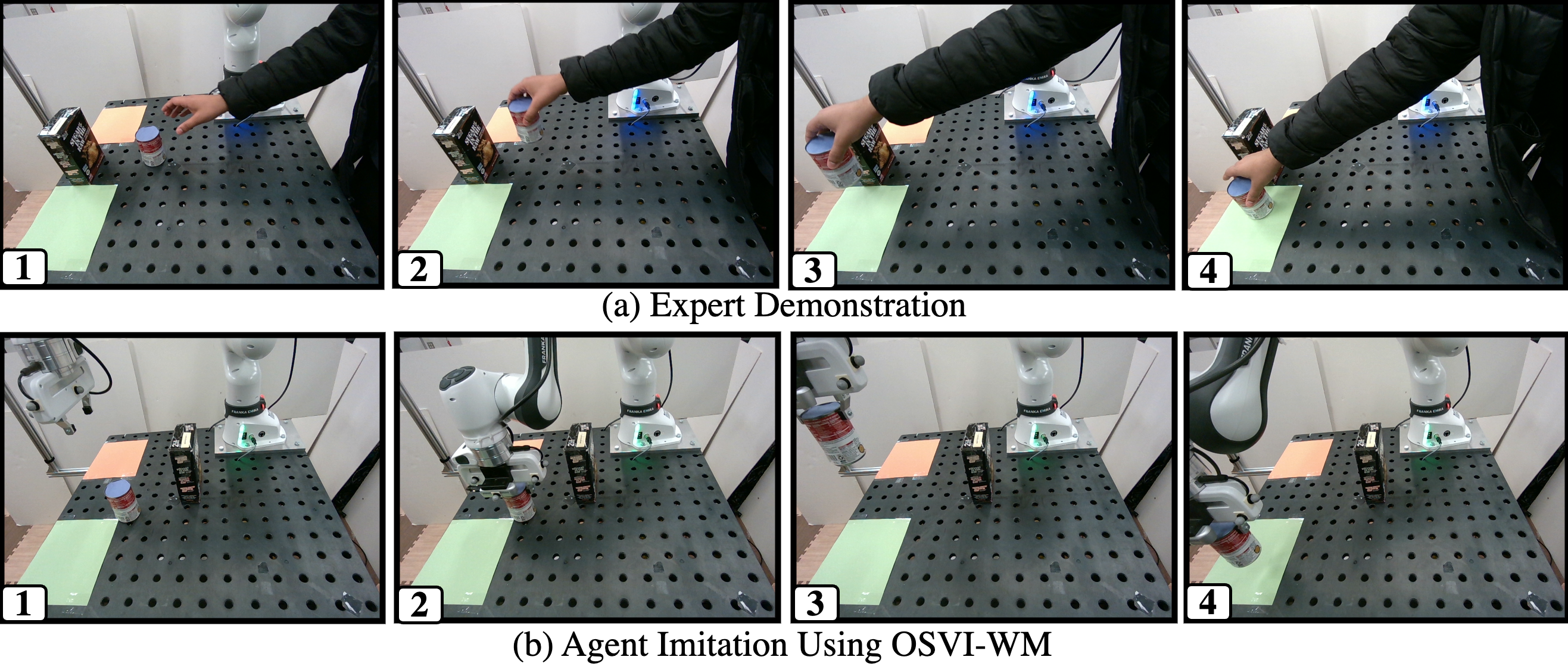}
    \caption{\textbf{Real-World Qualitative Example}: The human expert demonstrates the test task of picking and placing a can in the green target, which the agent successfully imitates using OSVI-WM.}
    \label{fig:qual_example}
    \vspace*{-0.5cm}
\end{figure}

We present a qualitative example of OSVI-WM in the Human-Franka-PP real-world setup. In Fig.~\ref{fig:qual_example}a, a human expert demonstrates the task of placing a can on a green target. Using only this demonstration, the agent (white Franka arm) successfully completes the task (Fig.~\ref{fig:qual_example}b), despite changes in object locations and embodiment differences. This task was excluded from training, illustrating OSVI-WM’s ability to generalize to unseen tasks by following the expert’s intent rather than relying solely on visual semantics. This example showcases OSVI-WM’s effectiveness under all key conditions outlined in our problem setup: unseen tasks, embodiment mismatch, unaligned trajectories, and in-domain training. More visualizations on different tasks are in the appendix.

\begin{figure}[t]
    \centering
    \begin{minipage}[t]{0.56\linewidth}
        \centering
        \includegraphics[trim={0 15 0 0},clip,width=\linewidth]{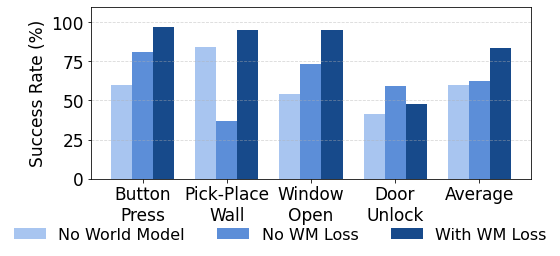}
        \caption{Ablation study on the world model (WM) and WM loss. Using both achieves the best performance.}
        \label{fig:ablation_wm}
    \end{minipage}
    \hfill
    \begin{minipage}[t]{0.39\linewidth}
        \centering
        \includegraphics[trim={0 15 0 0},clip,width=\linewidth]{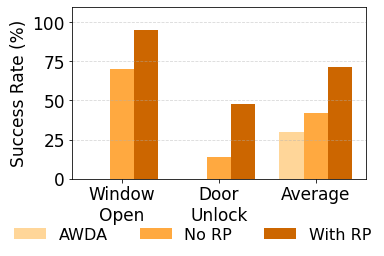}
        \caption{Ablation Study on OSVI-WM's Re-Planning (RP).}
        \label{fig:ablation_replanning}
    \end{minipage}
    \vspace*{-0.7cm}
\end{figure}

% \begin{table}[t]
%     \centering
%     \caption{Ablation results (average success rates) on the Pick Place~\cite{tosvi} and Meta-World~\cite{metaworld} benchmarks, evaluating the impact of different architectural choices by individually disabling them.}
%     \vspace{0.1cm}
%     % \setlength{\tabcolsep}{8pt}
%     % \renewcommand{\arraystretch}{1.4}
%     \begin{tabular}{ccccc}
%         \toprule
%         Stop Grad & Spatial Pool & Multi-Step & \multicolumn{2}{c}{\textbf{Benchmarks}} \\
%         in WM Loss & after WM & Prediction & Pick-Place & Meta-World \\
%         \midrule
%         $\times$ & \checkmark & \checkmark & 20 & 76.5 \\
%         \checkmark & $\times$ & \checkmark & 15 & 79.5 \\
%         \checkmark & \checkmark & $\times$ & \cellcolor{gray!20}\textbf{100} & 67.3 \\
%         \checkmark & \checkmark & \checkmark & \cellcolor{gray!20}\textbf{100} & \cellcolor{gray!20}\textbf{83.8} \\
%         \bottomrule
%     \end{tabular}
%     \label{tab:ablation}
%     \vspace*{-0.5cm}
% \end{table}

\noindent \textbf{What is the impact of the world-model-guided trajectory generation module?}
We assess the impact of this module through two ablations on the Meta-World dataset. The first removes the world model loss $\mathcal{L}_{wm}$; the second removes the world-model-guided trajectory generation module entirely, replacing it with a single action model that predicts an action vector from the expert demonstration and the agent’s initial observation. This vector is passed through the pooling and waypoint prediction modules. As shown in Fig.~\ref{fig:ablation_wm}, removing the world model yields the lowest performance, while adding it without $\mathcal{L}_{wm}$ provides limited gains. 
The loss $\mathcal{L}_{wm}$ is applied to future latent states predicted by the world model and is independent of ground-truth waypoint supervision. It acts as an auxiliary dynamics-consistency regularizer and encourages robust, task-agnostic representations, thus improving  generalization to unseen tasks.
Hence, combining the world model with $\mathcal{L}_{wm}$ achieves the best performance, underscoring the importance of both the module and its training signal.

\begin{figure}[t]
    \centering
    \includegraphics[width=.9\linewidth]{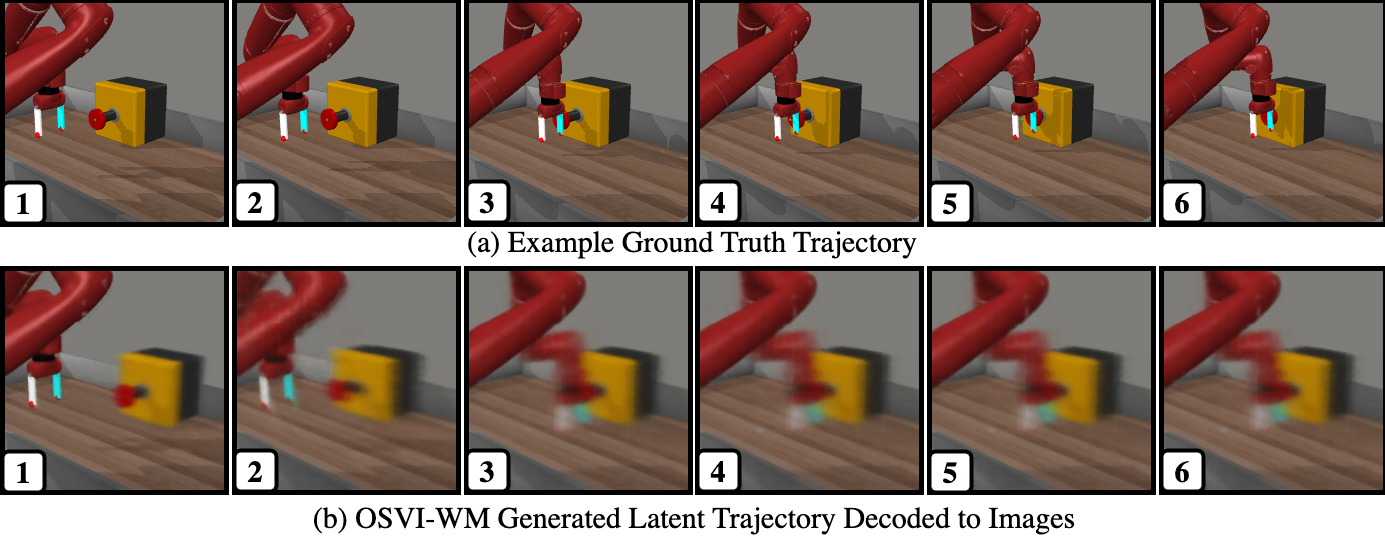}
    \caption{(a) An example ground truth rollout of the Button-Press-V2 task from Meta-World~\cite{metaworld}. (b) Latent trajectory generated using OSVI-WM's world model decoded into images for visualization. }
    \label{fig:predicted_traj}
    \vspace*{-0.7cm}
\end{figure}

In Fig.\ref{fig:predicted_traj}, we visualize the predicted trajectory generated by OSVI-WM’s world model on the Button-Press-V2 task~\cite{metaworld}. 
To convert the latent states into images, we train a Transpose Convolution-based image predictor. Images from the task are encoded using OSVI-WM’s trained encoder and spatial pooling layer, and the image predictor is trained to predict back the pixels. During evaluation, we generate the future latent states ($r_2,\dots,r_M$) using the expert demonstration and initial agent frame, and decode them into images (Fig.\ref{fig:predicted_traj}b) using the image predictor. A ground truth rollout is shown for comparison in Fig.~\ref{fig:predicted_traj}a.
Two key observations emerge: (a) the predicted trajectory captures meaningful, task-completing states, effectively summarizing the task; (b) the latent states are informative enough to reconstruct interpretable images. As expected, later states appear more pixelated due to accumulated errors in recursive latent state predictions. Despite this, the decoded trajectory clearly reflects the full task execution and produces accurate downstream performance.

\noindent \textbf{Which components of OSVI-WM are most critical to its success?}
As shown above, the world-model-guided trajectory generation module, along with supervision from the WM loss $\mathcal{L}_{wm}$, is critical to OSVI-WM’s success. Re-planning further boosts performance, particularly on the challenging Meta-World `hard' tasks. Fig.~\ref{fig:ablation_replanning} isolates the impact of re-planning, showing an approximate 30-point gain. 
For comparison, we also report the average success rate of AWDA~\cite{awda} (with additional training data). Notably, even without re-planning, OSVI-WM outperforms AWDA.
Table~\ref{tab:ablation} presents further ablation studies evaluating key architectural choices. Removing stop-gradient in the WM loss significantly degrades performance, dropping Meta-World accuracy by 7\% and Pick-and-Place by 80\%, clearly indicating model collapse. 
\begin{wraptable}{r}{0.63\columnwidth}
    \vspace{-0.3cm}
    \centering
    \caption{Ablation results (avg. success rates) on the Pick-Place~\cite{tosvi} and Meta-World~\cite{metaworld}, evaluating the impact of different architectural choices by individually disabling them.}
    % \vspace{0.05cm}
    \setlength{\tabcolsep}{1.5pt}
    \begin{tabular}{ccccc}
        \toprule
        Stop Grad & Spatial Pool & Multi-Step & \multicolumn{2}{c}{\textbf{Benchmarks}} \\
        in WM Loss & after WM & Prediction & Pick-Place & Meta-World \\
        \midrule
        $\times$ & \checkmark & \checkmark & 20 & 76.5 \\
        \checkmark & $\times$ & \checkmark & 15 & 79.5 \\
        \checkmark & \checkmark & $\times$ & \cellcolor{gray!20}\textbf{100} & 67.3 \\
        \checkmark & \checkmark & \checkmark & \cellcolor{gray!20}\textbf{100} & \cellcolor{gray!20}\textbf{83.8} \\
        \bottomrule
    \end{tabular}
    \label{tab:ablation}
    \vspace{-0.4cm}
\end{wraptable}
Applying spatial pooling before the world-model-guided trajectory generation block introduces an information bottleneck, resulting in a 85\% drop in Pick-and-Place and an 4\% drop in Meta-World. Switching from multi-step recursive to single-step trajectory prediction has no effect on Pick-and-Place but lowers Meta-World performance by over 15\% due to Meta-World's greater complexity.
Together, these results show that OSVI-WM’s effectiveness stems from key components: (a) the world-model-guided trajectory generation with WM loss, (b) re-planning, (c) stop-gradient in WM supervision, (d) spatial pooling after the world model, and (e) multi-step trajectory prediction. Regularization strategies like asymmetric demonstration image mixup~\cite{awda} further improve generalization.

\noindent \textbf{Division of train and test tasks:} 
The choice of train–test task division affects performance, especially in the Meta-World simulation environment. During training, the model learns two key components: inferring tasks from the expert and acquiring the skills needed to complete these tasks. Hence, careful consideration is required when choosing training tasks to ensure meaningful generalization to unseen tasks.
To further investigate this, we conducted two ablation studies on the Meta-World benchmark. In the first, we excluded all button-related tasks (Button-Press-Topdown, Button-Press-Topdown-Wall, Button-Press-Wall) from training and evaluated on Button-Press. The model still achieved a 97\% success rate, nearly identical to when the button-related tasks were included. This is because button pressing is relatively kinematically less complicated and the required skills were already acquired from other training tasks.
In the second ablation, we removed Window-Close from the training set and evaluated on Window-Open. Here, performance decreased from 95\% to 80\%.The sliding window tasks here involve more complex physical dynamics. So the absence of related training data led to a noticeable drop in generalization. If the training set included a very large variety of tasks with diverse motions and skills, the model would potentially perform well on any task, even if no similar tasks were present in the training set. However, this approach would require large out-of-context datasets, significant training resources, and extended training times. In OSVI-WM, we follow the train–test split proposed in AWDA~\cite{awda} for consistency with prior work.

\section{Conclusion}
\label{sec:conclusion}
\label{sec:limitations}
% \vspace*{-0.3cm}

We introduced OSVI-WM, a novel framework for one-shot visual imitation on unseen tasks. Given a single expert demonstration and the agent’s initial observation, OSVI-WM encodes them into latent states and uses a world-model-guided approach to predict a trajectory of future latent states. This trajectory is decoded into physical waypoints that control the robot. Extensive experiments in both simulation and real-world settings demonstrate OSVI-WM’s strong performance, and ablation studies confirm the importance of key design choices in the framework.
By generalizing effectively from a single demonstration to unseen tasks, OSVI-WM takes a step toward enabling autonomous robotic systems capable of operating in complex, real-world domains such as industrial automation and medical assistance, without constant human supervision.

\noindent \textbf{Limitations}: While OSVI-WM demonstrates promising results, there are limitations that open avenues for future work. Like prior work~\cite{tosvi, awda, mosaic}, it only predicts 3D positions and a gripper state, without modeling the end-effector’s orientation. This may limit its applicability to tasks requiring rotational actions, such as screwing or side-grasping. The current formulation focuses on single-task execution; extending the framework to sequential or multi-task settings is an interesting direction for future work. The division of train and test task can potentially influence the performance of OSVI-WM as detailed in Section~\ref{sec:results}. Further, although the test tasks are unseen during training, they belong to the same family of manipulators as the training data.
While these limitations do not hinder current performance, they highlight promising directions for enhancing the framework in more complex task settings.

\section{Acknowledgements}
This paper is supported in part by the Army Research Office under grant number W911NF21-1-0155, by the New York University Abu Dhabi (NYUAD) Center for Artificial Intelligence and Robotics, funded by Tamkeen under the NYUAD Research Institute Award CG010, and by NSF under grant number 2208189.

\bibliographystyle{plainnat}
\bibliography{main}  % .bib

%%%%%%%%%%%%%%%%%%%%%%%%%%%%%%%%%%%%%%%%%%%%%%%%%%%%%%%%%%%%

\newpage
\renewcommand{\thesection}{A\arabic{section}}
\renewcommand{\thefigure}{A\arabic{figure}}
\renewcommand{\thetable}{A\arabic{table}}
\setcounter{section}{0}  % Reset section counter
\setcounter{figure}{0}  % Reset figure counter
\setcounter{table}{0}  % Reset table counter
\section*{Appendix}
\section{Model Details}
\subsection{Encoder}
As described in Section 3 of the manuscript, we use a ResNet-18 encoder~\cite{resnet} to extract image features, producing a feature map of shape $(F, H, W)$ with $F=512$. For Meta-World~\cite{metaworld} images at a resolution of $(224, 224)$, this results in $(H, W) = (7, 7)$. For other benchmarks with input resolution $(240, 320)$, the feature map has dimensions $(H, W) = (8, 10)$. Following prior work~\cite{dynamo,tosvi,awda}, we initialize the ResNet-18 with ImageNet-pretrained weights to facilitate better convergence.

\subsection{Action and World Models}
At each timestep $i$, the expert states ${e_N, \dots, e_1}$ and the agent states ${r_i, \dots, r_1}$ are concatenated along the temporal dimension, resulting in a tensor of shape $(N+i, F \times H \times W)$. Sinusoidal positional embeddings are then added along the time dimension, and the sequence is processed by causal transformer blocks to produce the latent action $a_i$.
For Meta-World~\cite{metaworld}, we use 6 causal transformer blocks, whereas for other benchmarks, we use 2 blocks. All transformer blocks use 4 attention heads.
For the world model, the latent actions ${a_i, \dots, a_1}$ are concatenated with the corresponding states along the feature dimension and passed through a single causal transformer block with 4 attention heads. The output at the final timestep is projected to shape $(F \times H \times W)$ to predict next state $r_{i+1}$.

\subsection{Grasp Correction}
Following the approach in~\cite{awda}, we use a depth camera mounted on the robot’s end-effector to refine the grasp pose. Specifically, when the agent reaches a waypoint that requires closing the gripper (i.e., initiating a grasp), it uses the depth camera to detect all visible connected components and identifies the largest one. The end-effector is then adjusted so that this largest contour is positioned directly beneath it. In practice, this grasp correction step is repeated for three consecutive iterations each time a grasp is triggered.

\section{Training Details}
As mentioned in the manuscript, we train the model for 10 epochs on simulation benchmarks and 100 epochs on real-world benchmarks. Training is performed using the AdamW optimizer with a one-cycle learning rate scheduler and a batch size of 128. All models are implemented in PyTorch~\cite{pytorch} and trained using the accelerate library~\cite{accelerate} for multi-GPU parallelization across 4 NVIDIA A100 GPUs (40 GB each).
For evaluation, simulation benchmarks are run on an RTX A4000 GPU with 128 GB RAM and an Intel i9 CPU. Real-world benchmarks are evaluated on an RTX 2080 Ti GPU with 64 GB RAM and an Intel i7 CPU.

To improve generalization, we adopt the demonstration image mixup strategy, similar to prior work~\cite{awda}. During training, each input image $\mathcal{I}$ is linearly mixed with an image $\mathcal{J}$ from another task using a mixing factor $\beta \sim \mathcal{U}(0.3, 1.0)$ where $\mathcal{U}$ is the uniform distribution. The resulting input is $\bar{\mathcal{I}} = \beta \mathcal{I} + (1-\beta)\mathcal{J}$. This encourages the model to rely on task-relevant cues rather than image semantics.
Additionally, we apply regularization techniques such as random cropping, random translation, and color jittering to enhance robustness and performance.

\textbf{Loss Functions}: 
As described in the manuscript, we use the L1 loss for supervising the world model’s outputs and the differentiable Soft Dynamic Time Warping (Soft-DTW)~\cite{soft_dtw} loss for training the predicted waypoints. Dynamic Time Warping (DTW) is commonly used to measure the similarity between temporal sequences that may vary in length or speed, by solving a minimum-cost alignment between them using dynamic programming. However, the standard DTW loss is not differentiable with respect to its inputs, which limits its applicability in gradient-based optimization.
To address this, \cite{soft_dtw} introduced Soft-DTW, a differentiable relaxation of DTW that replaces the hard minimum operator in the dynamic programming recurrence with a soft-minimum (log-sum-exp) formulation. This soft assignment aggregates alignment costs across all possible paths, weighted by their relative scores, thereby allowing gradients to flow through all potential alignments. In our work, we adopt this differentiable Soft-DTW loss to effectively supervise the predicted waypoint sequences and enable end-to-end training.

\section{Benchmarks Environments}
\subsection{Meta-World}
The Meta-World benchmark~\cite{metaworld} comprises 50 diverse manipulation tasks performed using a Sawyer robotic arm. Following the setup in~\cite{awda}, we use 46 tasks for training and reserve 4 tasks for testing. For each of the 46 training tasks, we collect 100 trajectories from both the expert and the agent across varied, randomly sampled environment configurations. These trajectories are used for training the model. The test tasks are grouped into `easy' and `hard' categories.

The easy tasks include Pick-Place-Wall-V2 and Button-Press-V2. Pick-Place-Wall-V2 differs from Pick-Place-V2 from the training set by introducing a wall between the pick and place locations. As a result, the manipulator must lift the object higher to avoid collision, posing a non-trivial challenge. Similarly, Button-Press-V2 differs from Button-Press-Wall-V2 in the training task by omitting the wall, altering the spatial context.

The hard tasks are Window-Open-V2 and Door-Unlock-V2. Window-Open-V2 is semantically similar to Window-Close-V2 from the training set but requires performing the reverse action. Successfully completing this task requires the model to follow the expert demonstration rather than relying solely on scene semantics. Likewise, Door-Unlock-V2 is the reverse of the training task Door-Lock-V2, also demanding fine-grained control and precision, making it particularly challenging.

\subsection{Pick-and-Place}
The Pick-and-Place benchmark, introduced in T-OSVI~\cite{tosvi}, features a tabletop environment with four distinct objects, varying in size and shape, placed on one side, and four target locations on the opposite side. This setup defines 16 unique pick-and-place pairs, each representing a separate task within the benchmark. Following~\cite{awda}, we use 14 tasks for training and reserve 2 task for testing.

A key challenge in this benchmark is the embodiment mismatch, where the expert demonstrations are performed using a Sawyer robotic arm, while the agent executes tasks using a Franka arm.
As in the Meta-World setup, we collect 100 trajectories for each training task from both the expert and the agent under randomly sampled environment configurations, which are used to train the model.

\subsection{Two-Franka-PP}
This benchmark involves two Franka robotic arms mounted at different locations on a shared tabletop environment, as illustrated in Fig. 5a of the manuscript. The gray Franka serves as the expert, while the white Franka acts as the agent. We utilize an Intel RealSense D455 camera with known intrinsic and extrinsic (camera-to-robot transformation) parameters, capturing images at 640x480 resolution. Additionally, an Intel RealSense D435 is mounted on the robot's end-effector to assist with grasping, as detailed in Section 3.2 of the manuscript. We use these same set of cameras for Human-Franka-PP and Human-Franka-Push benchmarks as well. The task in Two-Franka-PP is a pick-and-place operation involving two objects: a soup can and a sugar box. The corresponding target locations are positioned near the center of the table, marked with green and orange paper, respectively.
Due to the high cost of real-world data collection, we sample 30 trajectories, with different configurations of the environment, each for the expert and the agent across four tasks. Following the same train-test split protocol as prior benchmarks, we use 3 tasks for training and 1 for testing.
This real-world setup introduces key challenges like domain shift arising from the differing mount positions of the expert and agent arms, and the limited number of trajectories available for training.

\subsection{Human-Franka-PP}
This benchmark adopts a similar pick-and-place setup as Two-Franka-PP, but replaces the expert with a human demonstrator, while the agent remains a Franka robotic arm, as shown in Fig. 5b of the manuscript. In addition to the change in embodiment, the position and orientation of the external camera, as well as the locations of the target areas, are modified in this setting.
As in the previous real-world benchmark, we collect 30 trajectories per task with different configurations of the environment. We use 3 tasks for training and reserve 1 for testing.
The high embodiment gap between the human expert and robotic agent, combined with the limited number of trajectories, makes Human-Franka-PP a particularly challenging benchmark. 

\subsection{Human-Franka-Push}
This benchmark features a tabletop environment with a blue and a green cuboid-shaped objects. The objective is to either push or pull one of the objects, resulting in a total of four distinct tasks. We use 3 tasks for training and 1 for testing.
In this setup, the expert is a human demonstrator, while the agent is a Franka robotic arm. As with the other real-world benchmarks, we collect 30 trajectories each from the expert and the agent, using varied configurations of the environment to promote generalization.
Beyond the embodiment gap between the human and robotic agent, this benchmark poses an additional challenge: the required manipulations demand highly precise, fine-grained control. The low margin for error in object interactions makes accurate trajectory execution critical for successful task completion.

\section{Additional Comparisons}
\subsection{Visualization Examples}
\begin{figure}[H]
    \centering
    \begin{subfigure}{\linewidth}
        \centering
        \includegraphics[width=.9\linewidth]{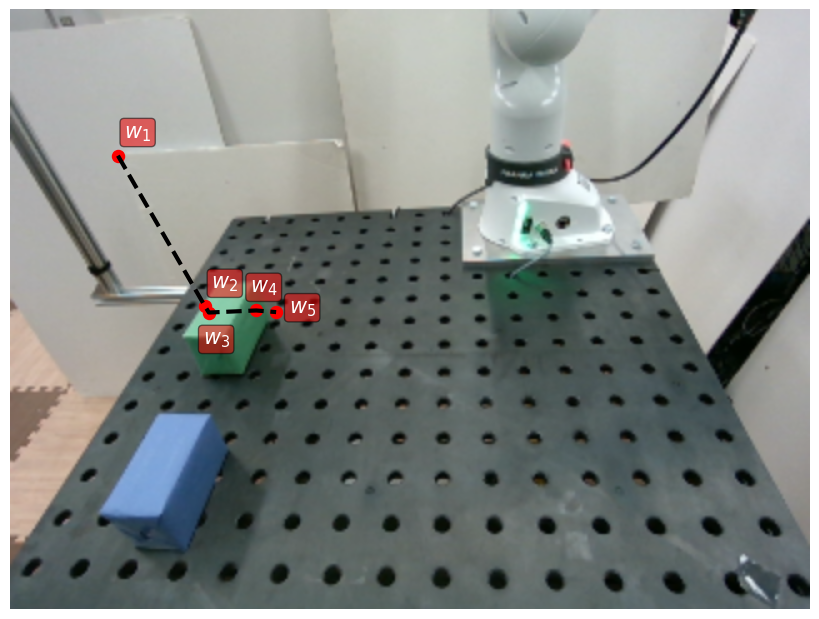}
        \caption{Waypoints predicted by AWDA~\cite{awda}.}
        \label{fig:waypoints_awda}
    \end{subfigure}
    % \vspace{0.5em} % optional spacing between subfigures
    \begin{subfigure}{\linewidth}
        \centering
        \includegraphics[width=.9\linewidth]{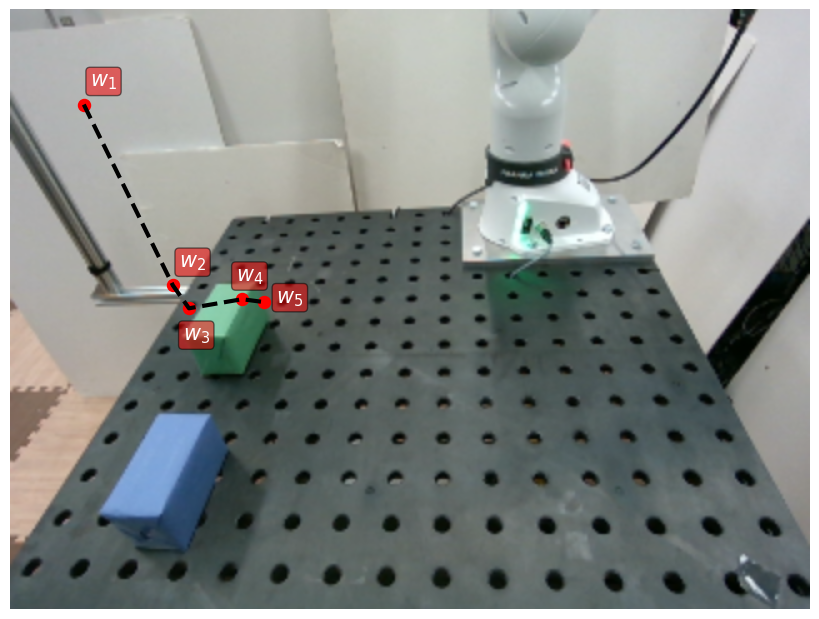}
        \caption{Waypoints predicted by OSVI-WM.}
        \label{fig:waypoints_ours}
    \end{subfigure}
    
    \caption{Comparison of predicted waypoints on a Human-Franka-Push example. The waypoints predicted by AWDA, particularly $w_2$ and $w_3$, cause the robot to collide with the object due to imprecise trajectory planning. In contrast, the waypoints generated by OSVI-WM are more accurate and enable smoother manipulation.}
    \label{fig:waypoints_comparison}
\end{figure}

\noindent \textbf{Waypoint Comparisons}
In this section, we compare the waypoints predicted by OSVI-WM (Fig.\ref{fig:waypoints_ours}) and AWDA\cite{awda} (Fig.~\ref{fig:waypoints_awda}) on an example from the Human-Franka-Push benchmark. This particular example was selected because it comes from a real-world setting and requires fine-grained control for successful execution. As shown in the figure, the waypoints predicted by AWDA, particularly $w_2$ and $w_3$, cause the robot to collide with the object due to imprecise trajectory planning. In contrast, the waypoints generated by OSVI-WM are more accurate and better aligned with the task requirements, enabling smoother manipulation and resulting in higher success rates.

\begin{figure}[H]
    \centering
    \includegraphics[width=\linewidth]{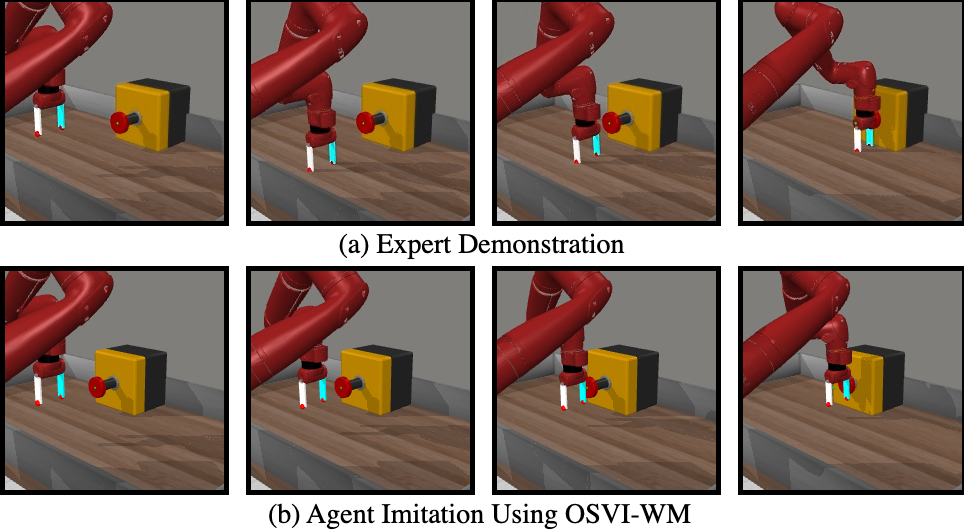}
    \caption{\textbf{Meta-World~\cite{metaworld} Button-Press-V2}: In this test task, the button is positioned at different locations across the expert and agent runs.}
    \label{fig:qual_button_pres}
\end{figure}

\begin{figure}[H]
    \centering
    \includegraphics[width=\linewidth]{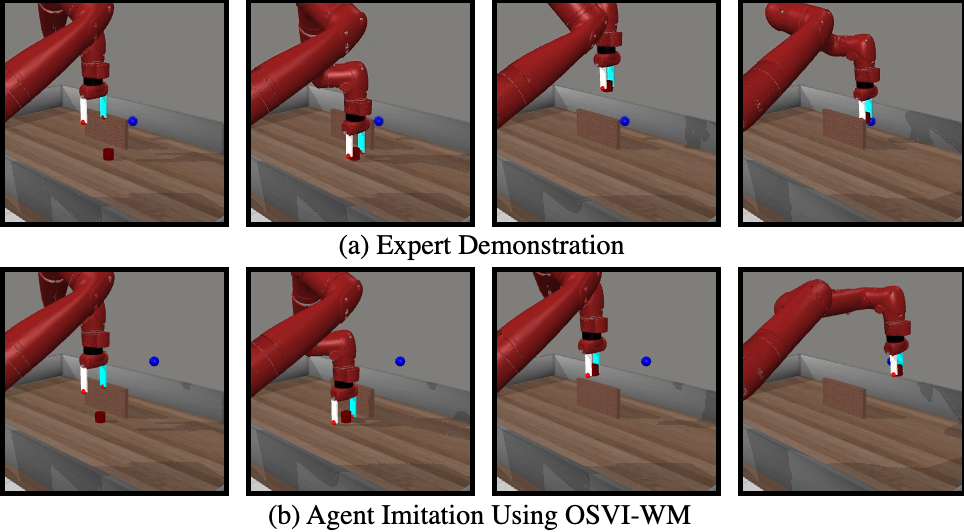}
    \caption{\textbf{Meta-World~\cite{metaworld} Pick-Place-Wall-V2}: In this test task, the object's initial location, target location (blue dot), and location of the wall are different across the expert and agent runs.}
    \label{fig:qual_pick_place_wall}
\end{figure}

\begin{figure}[H]
    \centering
    \includegraphics[width=\linewidth]{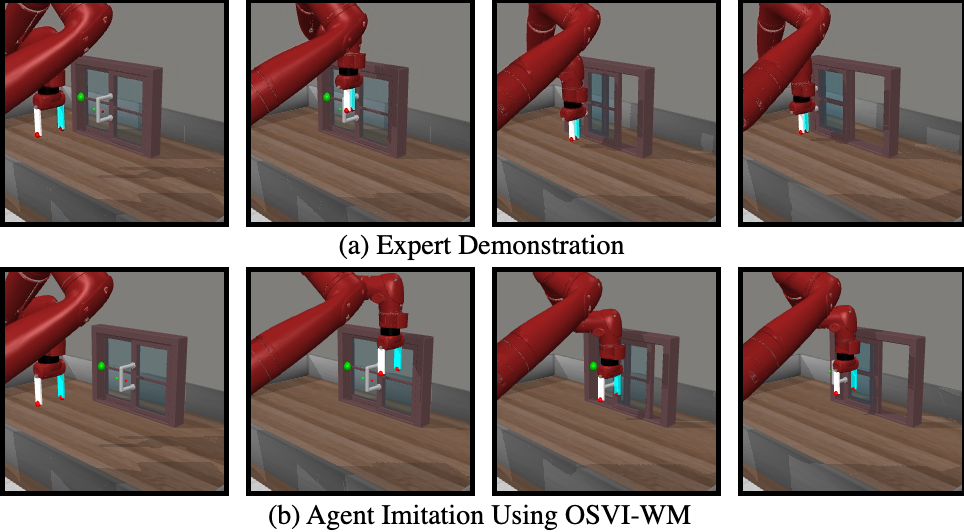}
    \caption{\textbf{Meta-World~\cite{metaworld} Window-Open-V2}: In this test task, the window is positioned at different locations between the expert and agent runs.}
    \label{fig:qual_window_open}
\end{figure}

\begin{figure}[H]
    \centering
    \includegraphics[width=\linewidth]{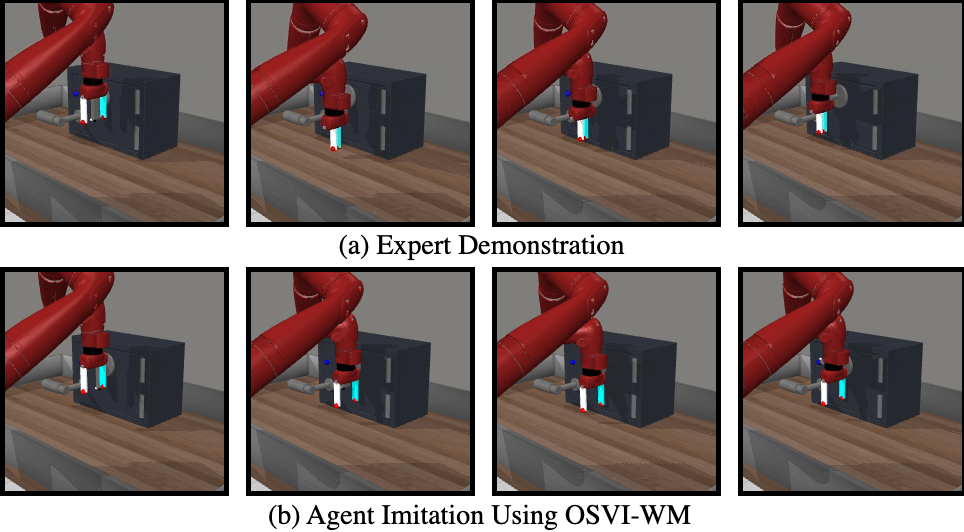}
    \caption{\textbf{Meta-World~\cite{metaworld} Door-Unlock-V2}: In this test task, the locker (with its door) is positioned at different locations between the expert and agent runs. Further, this task requires finer controls compared to other tasks, making it particularly challenging.}
    \label{fig:qual_door_unlock}
\end{figure}

\begin{figure}[H]
    \centering
    \includegraphics[width=\linewidth]{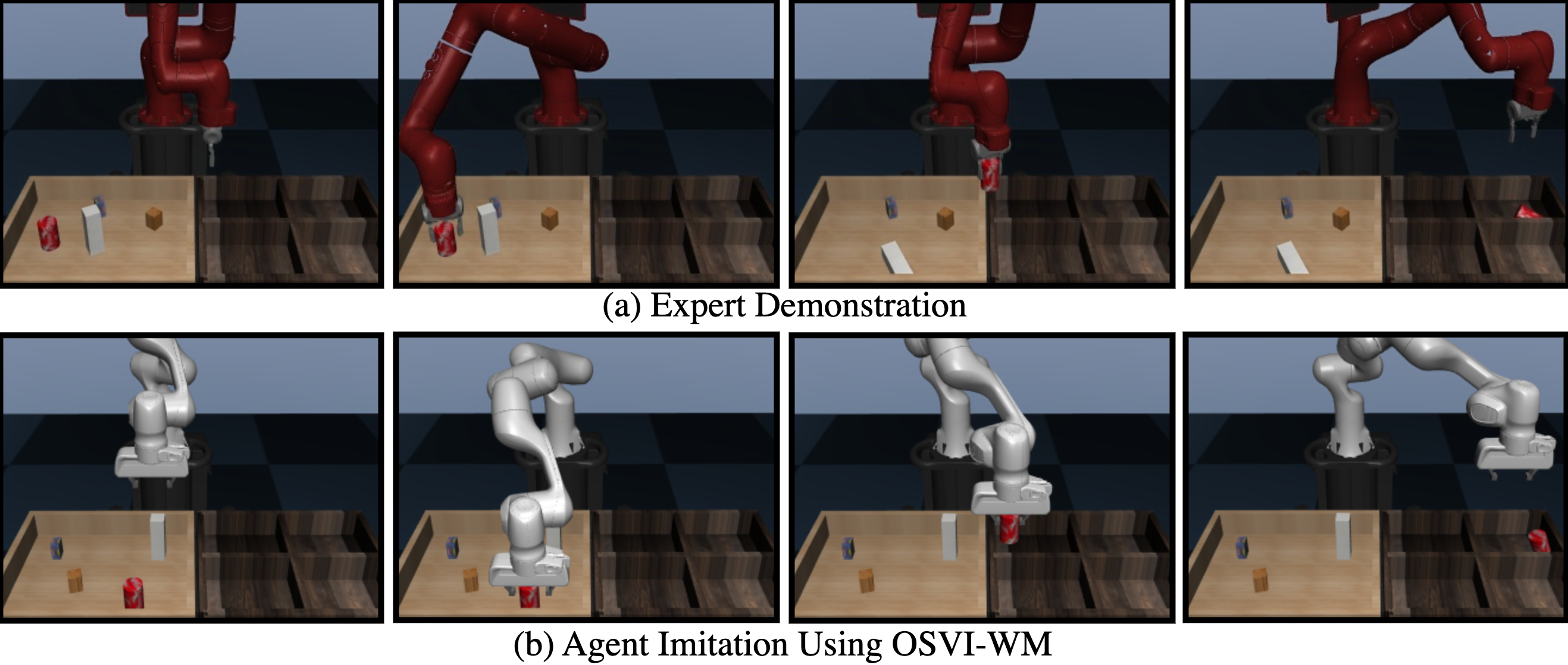}
    \caption{\textbf{Pick-and-Place~\cite{tosvi}}: The test task involves picking up the red can and placing it in the target location as demonstrated by the expert. The expert uses a Sawyer arm while the agent uses a Franka arm in this setting. Further, the arrangement of the objects is different between the expert demonstration and the agent's imitation rollout.}
    \label{fig:qual_tosvi}
\end{figure}

\begin{figure}[H]
    \centering
    \includegraphics[width=\linewidth]{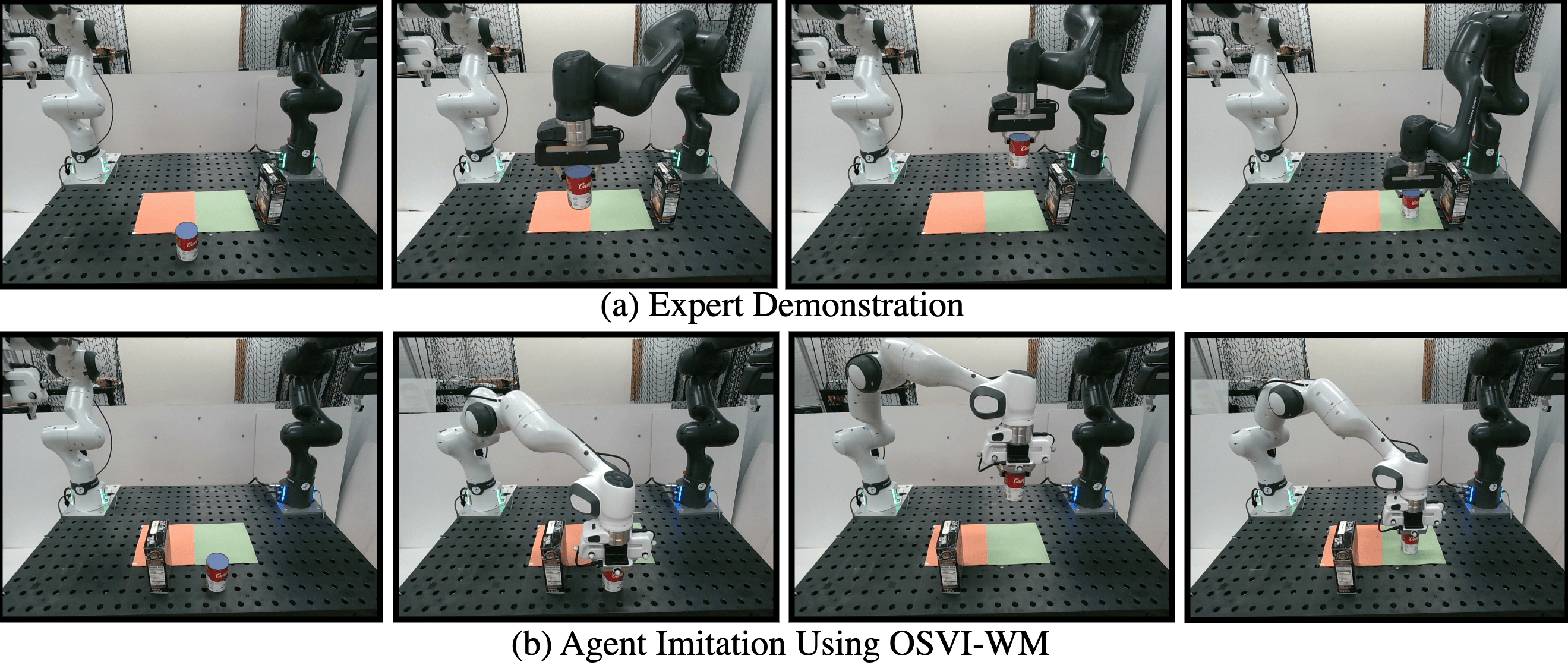}
    \caption{\textbf{Two-Franka-PP}: The goal is to pick up the can and place it into the green target, as demonstrated by the expert. The expert (gray Franka arm) and the agent (white Franka arm) are mounted at different locations on the tabletop. Additionally, the object arrangements differ between the expert demonstration and the agent’s imitation rollout.}
    \label{fig:qual_two_franka_pp}
\end{figure}

\begin{figure}[H]
    \centering
    \includegraphics[width=\linewidth]{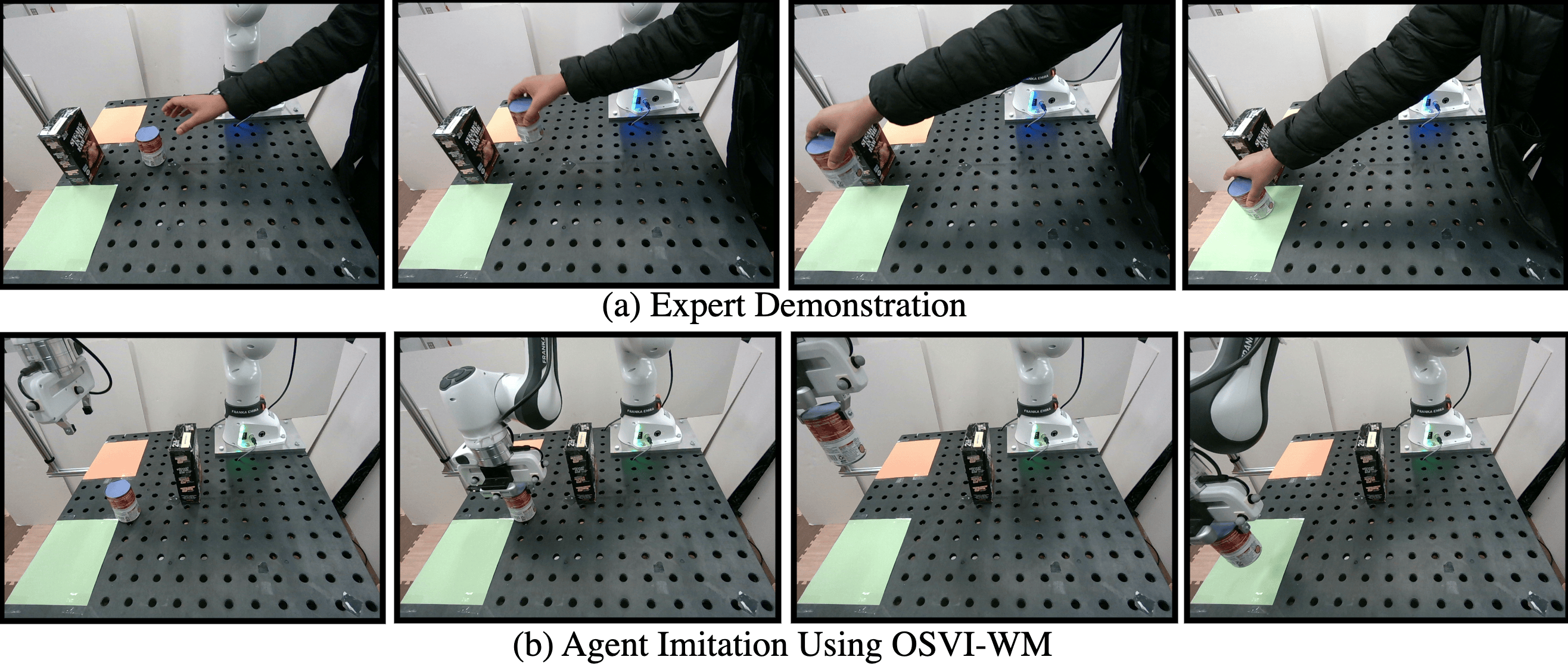}
    \caption{\textbf{Human-Franka-PP}: The goal is to pick up the can and place it into the green target, as demonstrated by the expert. The expert is a human arm, while the agent is a white Franka arm mounted on a tabletop. Additionally, the object arrangements differ between the expert demonstration and the agent’s imitation rollout.}
    \label{fig:qual_human_franka_pp}
\end{figure}

\begin{figure}[H]
    \centering
    \includegraphics[width=\linewidth]{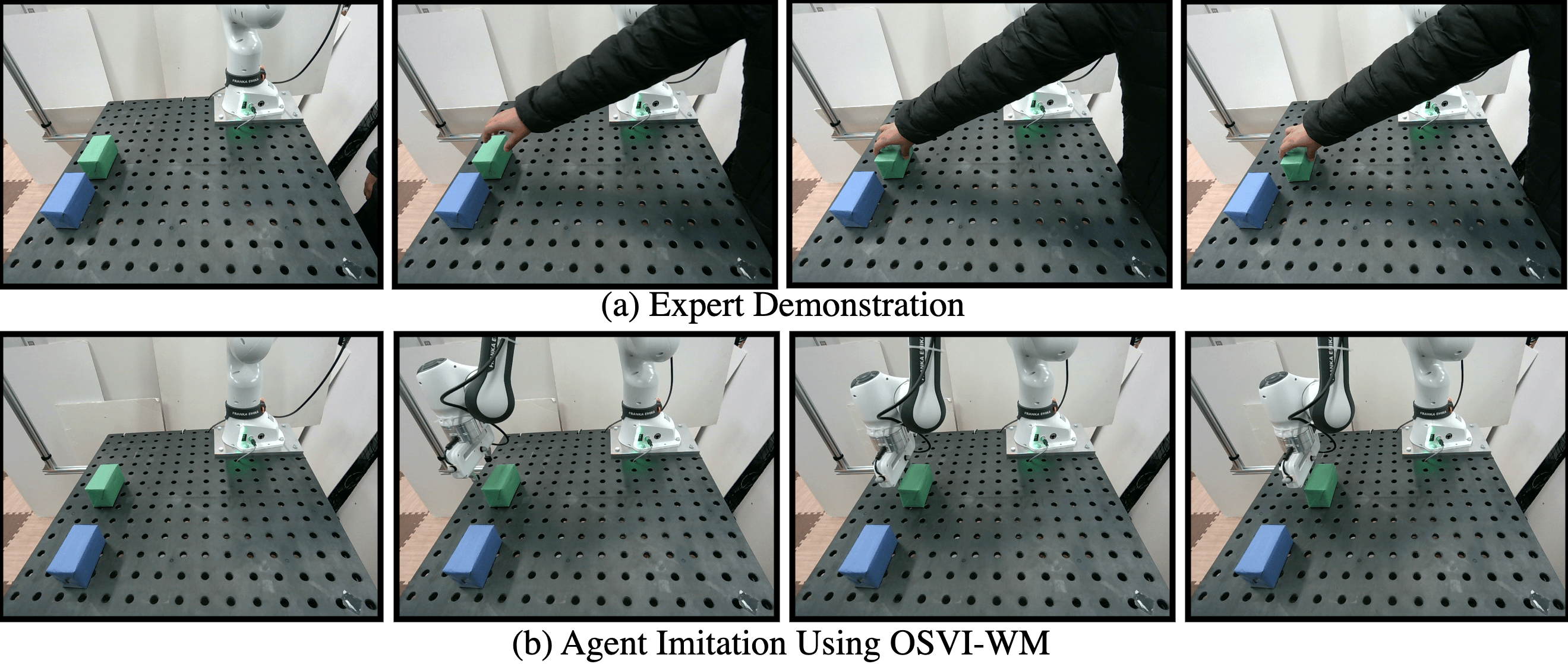}
    \caption{\textbf{Human-Franka-Push}: The objective is to push the green cuboid towards the agent, as shown in the figure. The expert is a human arm, while the agent is a white Franka arm mounted on a tabletop. Additionally, the object arrangements differ between the expert demonstration and the agent’s imitation rollout.}
    \label{fig:qual_human_franka_push}
\end{figure}

\begin{figure}[H]
    \centering
    \includegraphics[width=\linewidth]{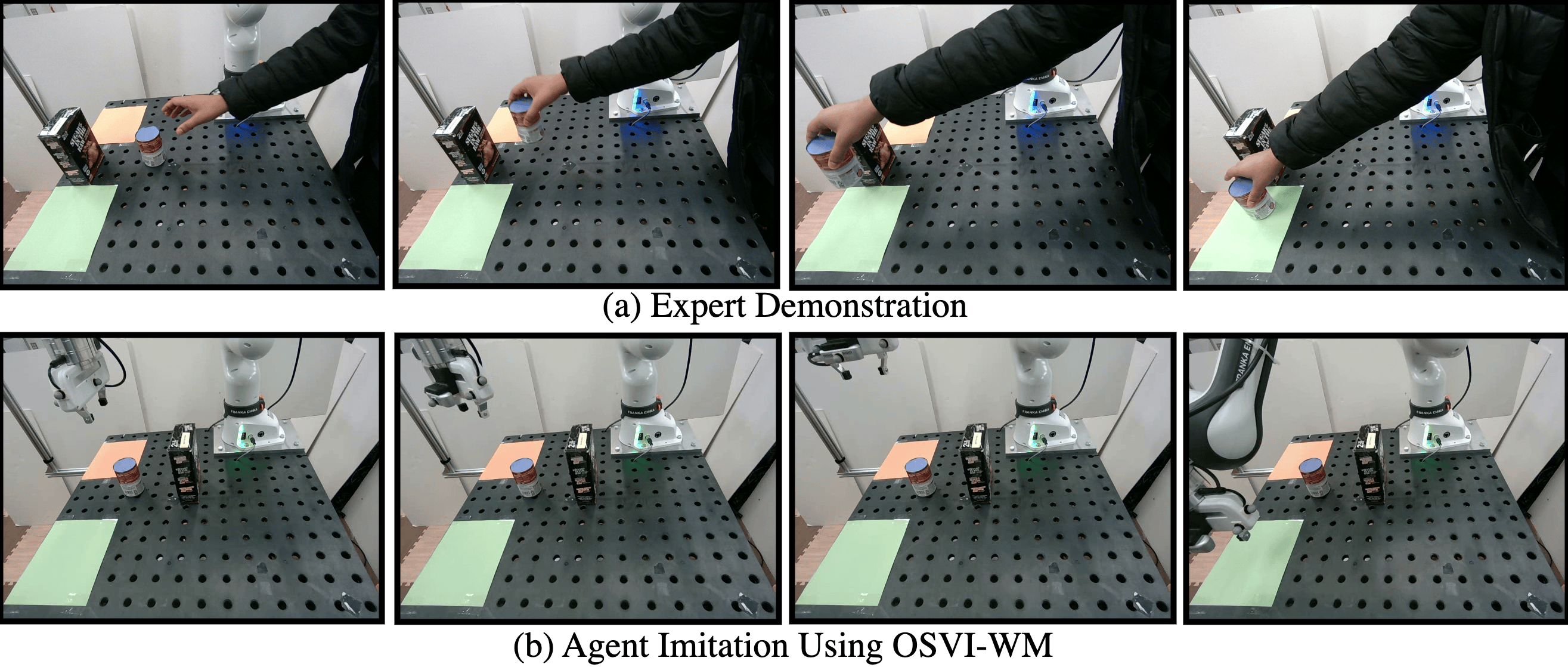}
    \caption{\textbf{Failure Case}: This example from the Human-Franka-PP benchmark illustrates a failure case. The agent, aiming to place the can on the green target, initially positions itself above the can but is misaligned by a few centimeters. As a result, it fails to establish a reliable grasp using the end-effector’s depth camera. The agent then skips the grasping stage and proceeds to the next waypoint without the object, ultimately reaching the target empty-handed and failing the task.}
    \label{fig:failure_human_franka_pp}
\end{figure}

\begin{figure}[H]
    \centering
    \includegraphics[width=\linewidth]{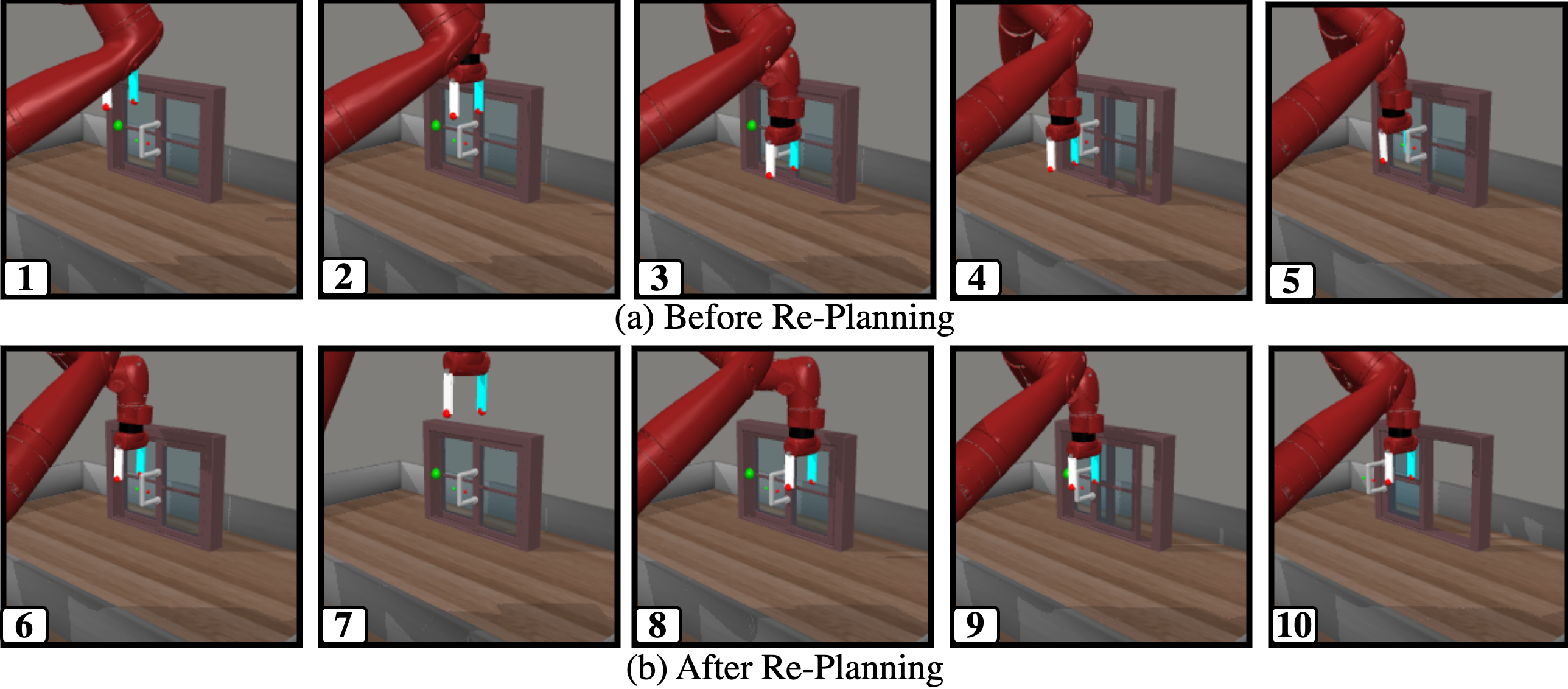}
    \caption{\textbf{OSVI-WM's Re-Planning Example}: The figure shows the agent’s rollout on the Window-Open-V2 test task from Meta-World~\cite{metaworld}. In steps 1–5 (a), the agent fails using the initial waypoints. Re-planning is triggered at step 5, enabling successful task completion in steps 6–10 (b).}
    \label{fig:qual_replanning}
\end{figure}

\noindent \textbf{Agent Rollouts}
We show visualization examples for each test task of Meta-World~\cite{metaworld} in Figs.~\ref{fig:qual_button_pres}, \ref{fig:qual_pick_place_wall}, \ref{fig:qual_window_open}, and \ref{fig:qual_door_unlock}, for Pick-and-Place~\cite{tosvi} in Fig.~\ref{fig:qual_tosvi}, and for the real-world benchmarks in Figs. \ref{fig:qual_two_franka_pp}, \ref{fig:qual_human_franka_pp}, and \ref{fig:qual_human_franka_push}. Further, we show a failure case in Fig. \ref{fig:failure_human_franka_pp}, and the use of replanning in Fig. \ref{fig:qual_replanning}. Details on the rollouts are provided in the respective captions.

\subsection{High-Level Comparison of World Models and VLAs for Robot Manipulation }
In general, both world-model-based and VLA-based (or behavior cloning) approaches ultimately aim to execute tasks by predicting actions. However, the key distinction is that world-model-based approaches also learn the world dynamics independently of the task. This discourages the model from overly fixating on action prediction and encourages learning more generic features through the additional training signal provided by the world model loss. Most existing methods, whether world-model-based or VLA-based, typically focus on training and testing within the same set of tasks, as detailed in our manuscript. Many of these methods have demonstrated good results in complex dexterous manipulation tasks, as you mentioned. However, when faced with out-of-distribution tasks, like those in our examples, existing BC-based methods such as~\cite{awda, tosvi, daml} often show low success rates, even on top-down grasping tasks. Additionally, our experience indicates that even recent VLA-based robot foundation models like Gr00t~\cite{groot} require in-context retraining for new tasks for accurate manipulation, despite being pre-trained on large datasets.

\subsection{Efficiency of Re-Planning}
When task execution fails (detected by the simulation environment configuration), OSVI-WM triggers re-planning by taking a new observation and generating a fresh sequence of waypoints based on the robot's current state. Unlike open-loop retries, replanning is a closed-loop process that takes a new observation of the current robot state and generates fresh waypoints based on that feedback. This closed-loop replanning incurs additional time cost. For the results shown in Figure 8, the time comparison is as follows:
\begin{itemize}
    \item Without re-planning: Each episode runs for a maximum of 500 steps (~40 seconds on our system with one RTX A4000 GPU, 128 GB RAM, and an Intel i9 CPU). \item With re-planning: New waypoints are generated every 500 steps, for up to 2000 steps (~160 seconds) total.
\end{itemize}

Despite this overhead, replanning substantially improves success rates. Notably, in our work, replanning is only applied for the `hard' Meta-World tasks. Even without replanning, OSVI-WM achieves a success rate of ~42\% on these tasks (71.5\% after re-planning), which is already higher than all existing baselines (Fig. 8). Additionally, it is important to highlight that many baseline methods, such as~\cite{tosvi}, are also inherently closed-loop but still perform worse than OSVI-WM, further emphasizing the strength of our approach beyond just repeated planning.

\raggedbottom

\end{document}